\documentclass{article}

    \usepackage[final,nonatbib]{neurips_2021}

\usepackage[square,sort,comma,numbers]{natbib}

\usepackage[utf8]{inputenc} 
\usepackage[T1]{fontenc}    
\usepackage{xspace}
\usepackage{url}            
\usepackage{booktabs}       
\usepackage{amsfonts}       
\usepackage{nicefrac}       
\usepackage{microtype}      
\usepackage{xcolor}         
 \usepackage{tikz}
\usepackage{mathtools}
\usepackage{amsmath}
\usepackage{amsthm}
\usepackage{booktabs}
\usepackage{floatrow}  
\usepackage{etoolbox}

\usepackage[capitalise,nameinlink]{cleveref}
\usepackage{amssymb,amsmath,amsthm}
\usepackage{centernot}
\usepackage{caption}
\usepackage{enumitem}
\usepackage{xcolor}
\usepackage{longtable}
\usepackage{wrapfig}
\usepackage{subcaption}
\usepackage{adjustbox}
\usepackage{algorithm}
\usepackage{algorithmic}
\usepackage{thmtools} 
\usepackage{thm-restate}
\usepackage{lipsum}
\usepackage{multirow}

\usepackage{bmpsize}

\DeclareMathOperator*{\argmax}{arg\,max}

\title{Risk-averse Heteroscedastic Bayesian Optimization}

\author{
        \hspace{-1em} Anastasiia Makarova \\
        \hspace{-1em} ETH Z{ü}rich \\
        \hspace{-1em} \texttt{anmakaro@ethz.ch} \\ 
    \And 
        \hspace{-1em} Ilnura Usmanova \\
        \hspace{-1em} ETH Z{ü}rich \\
        \hspace{-1em} \texttt{ilnurau@ethz.ch}\\ 
  \And 
      \hspace{-1em} Ilija Bogunovic \\
      \hspace{-1em} ETH Z{ü}rich \\
      \hspace{-1em}\texttt{ilijab@ethz.ch} \\
  \And 
      \hspace{-1em} Andreas Krause \\
      \hspace{-1em} ETH Z{ü}rich \\
      \hspace{-1em} \texttt{krausea@ethz.ch} \\
}

\begin{document}

\maketitle
\newcommand{\N}{\mathbb{N}}
\newcommand{\R}{\mathbb{R}}
\newcommand{\E}{\mathbb{E}}

\newcommand{\set}[1]{\mathcal #1}
\newcommand{\mat}[1]{\mathbf #1}
\renewcommand{\vec}[1]{\mathbf #1}
\newcommand{\x}{\mathbf{x}}
\newcommand{\w}{\mathbf{w}}
\newcommand{\X}{\mathcal{X}}
\newcommand{\D}{\mathcal{D}}
\newcommand{\ke}{\mathbf{k}}
\newcommand{\K}{\mathbf{K}}
\newcommand{\A}{\mathbf{A}}
\newcommand{\Y}{\mathbf{Y}}
\newcommand{\Si}{\Sigma}

\newcommand{\lm}{\boldsymbol{\lambda}}
\newcommand{\tht}{\boldsymbol{\theta}}
\newcommand{\ph}{\boldsymbol{\phi}}
\newcommand{\omg}{\boldsymbol{\omega}}

\newcommand{\todo}[1] {\textcolor{red}{#1}}
\newcommand{\todoformula}[1] {\mathbin{\textcolor{red}{#1}}}
\newcommand{\formulacomment}[1] {\mathbin{\textcolor{gray}{/#1/}}}
\newcommand{\iu}[1]{\textcolor{blue}{IU: #1}}
\newcommand{\ib}[1]{\textcolor{blue}{IB: #1}}

\newtheorem{theorem}{Theorem}
\newtheorem{corollary}{Corollary}[theorem]
\newtheorem{lemma}[theorem]{Lemma}
\newtheorem{prop}{Proposition}
\newtheorem{definition}{Definition}

\newcommand{\krepeat}{k}
\newcommand{\samplevar}{\hat{s}^2_{\krepeat}(x)}
\newcommand{\varproxy}{var}
\newcommand{\raucb}{\textsc{RAHBO}\xspace}
\newcommand{\gpucb}{\textsc{GP-UCB}\xspace}
\newcommand{\raucbus}{\textsc{RAHBO-US}\xspace}
\newcommand{\swissfel}{SwissFEL\xspace}
\newcommand{\xreported}{\hat{x}_T}
\newcommand{\mv}{\text{MV}\xspace}
\newcommand{\lcb}{\mathrm{lcb}\xspace}
\newcommand{\ucb}{\mathrm{ucb}\xspace}
\newcommand{\ubar}[1]{\text{\b{$#1$}}}
\newcommand{\rhomin}{\smash{\underline{\varrho}}}
\newcommand{\rhomax}{\bar{\varrho}}

\newcommand{\squeezeup}{\vspace{-2.5mm}}

\definecolor{aliceblue}{RGB}{0.94, 0.97, 1.0}

\newlength{\subcolumnwidth}
\newenvironment{subcolumns}[1][0.45\columnwidth]
 {\valign\bgroup\hsize=#1\setlength{\subcolumnwidth}{\hsize}\vfil##\vfil\cr}
 {\crcr\egroup}
\newcommand{\nextsubcolumn}[1][]{%
  \cr\noalign{\hfill}
  \if\relax\detokenize{#1}\relax\else\hsize=#1\setlength{\subcolumnwidth}{\hsize}\fi
}
\newcommand{\nextsubfigure}{\vfill}

\crefname{assumption}{assumption}{assumptions}
\Crefname{assumption}{Assumption}{Assumptions}
\newtheorem{assumption}{Assumption}

\newtheorem{innercustomthm}{Theorem}
\newenvironment{customthm}[1]
  {\renewcommand\theinnercustomthm{#1}\innercustomthm}
  {\endinnercustomthm}
  
\newcommand*\circled[1]{\tikz[baseline=(char.base)]{
            \node[shape=circle,draw,inner sep=2pt] (char) {#1};}}

\begin{abstract}
Many black-box optimization tasks arising in high-stakes applications require risk-averse decisions. The standard Bayesian optimization (BO) paradigm, however, optimizes the expected value only. We generalize BO to {\em trade mean and input-dependent variance} of the objective, both of which we assume to be unknown a~priori.
In particular, we propose a novel risk-averse heteroscedastic Bayesian optimization algorithm
(\raucb)  that aims to identify a solution with high return and low noise variance, while learning the noise distribution on the fly. To this end, we model both expectation and variance as (unknown) RKHS functions, and propose a novel risk-aware acquisition function. We bound the regret for our approach and provide a robust 
rule to report the final decision point
for applications where only a single solution must be identified. We demonstrate the effectiveness of \raucb on synthetic benchmark functions and hyperparameter tuning tasks. 
\end{abstract}

\section{Introduction}
\label{sec:into}

Black-box optimization tasks arise frequently in high-stakes applications such as drug and material discovery \cite{korovina2019chembo,griffiths2020,Negoescu2011}, genetics \cite{gonzalez2015bayesian,moss2020}, robotics \cite{berkenkamp2020bayesian,calandra2016,Marco_2016},
hyperparameter tuning of complex learning systems \cite{kirschner19a, chen2018bayesian,snoek2012practical}, to name a few. 
In many of these applications, there is often a trade-off between achieving high utility and minimizing risk. Moreover, uncertain and costly evaluations are an inherent part of black-box optimization tasks, and modern learning methods need to handle these aspects when balancing between the previous two objectives. 

Bayesian optimization (BO) is a powerful framework for optimizing such costly black-box functions from noisy zeroth-order evaluations. Classical BO approaches are typically \emph{risk-neutral} as they seek to optimize the expected function value only. 
In practice, however, two different solutions might attain similar expected function values, but one might produce significantly noisier realizations. This is of major importance when it comes to actual deployment of the found solutions. For example, when selecting hyperparameters of a machine learning algorithm, we might prefer configurations that lead to slightly higher test errors but at the same time lead to smaller variance. 

In this paper, we generalize BO to trade off mean and input-dependent noise variance when sequentially querying points and outputting final solutions. We introduce a practical setting where {\em both} the black-box objective and input-dependent noise variance are {\em unknown} a priori, and the learner needs to estimate them on the fly. We propose a novel optimistic risk-averse algorithm -- \raucb\ -- that makes sequential decisions by simultaneously balancing between \emph{exploration} (learning about uncertain actions), \emph{exploitation} (choosing actions that lead to high gains) and \emph{risk} (avoiding unreliable actions). We bound the cumulative regret of \raucb as well as the number of samples required to output a single near-optimal risk-averse solution. In our experiments, we demonstrate the risk-averse performance of our algorithm and show that standard BO methods can severely fail in applications where reliability of the reported solutions is of utmost importance.

\begin{figure}[t!]
    \centering
    \vspace{0.6cm}
    \hspace*{-1.5em}
    \subfloat[Unknown objective $f$]{
        \includegraphics[width=0.33\linewidth]{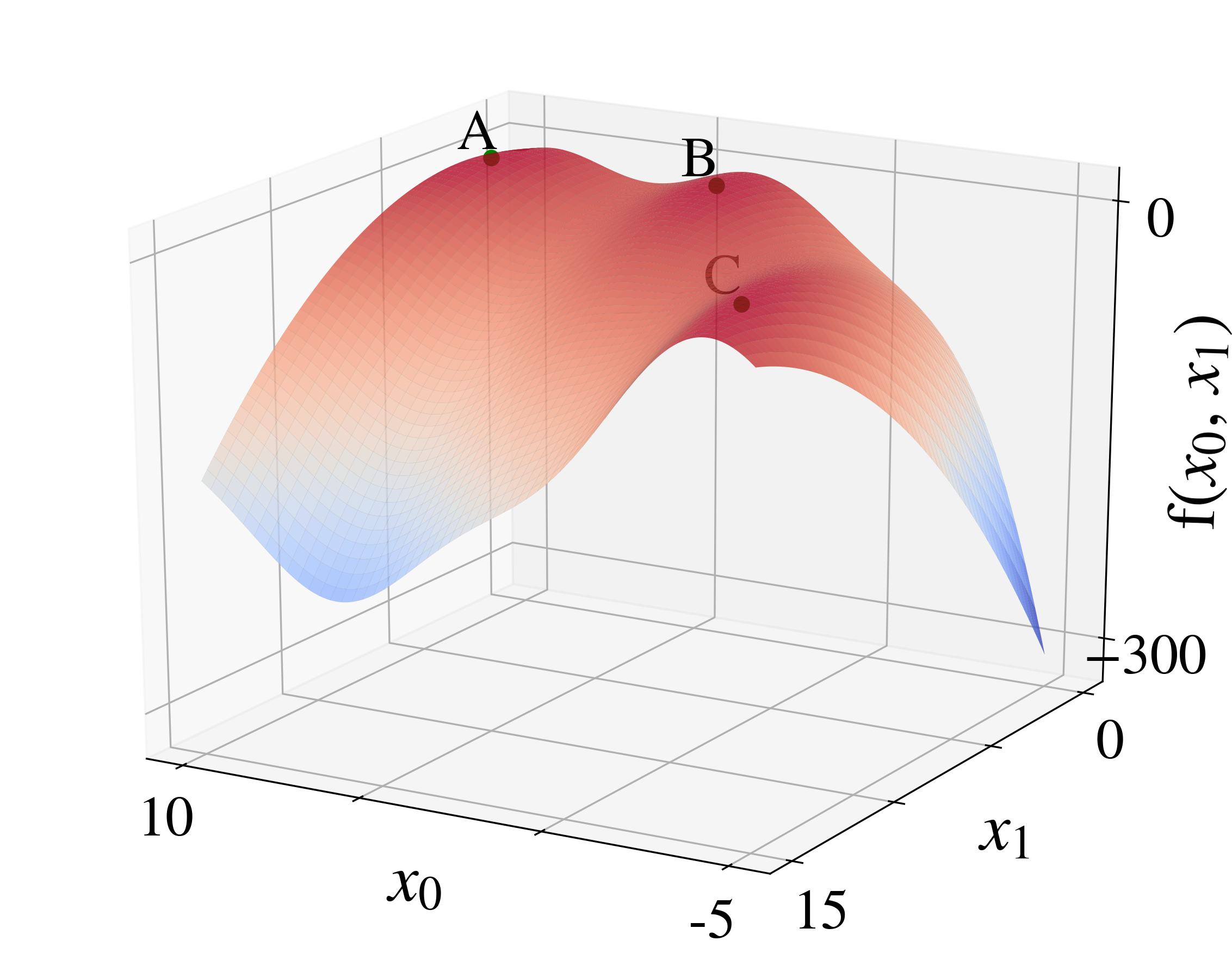}
        \label{fig:branin}
        }
    \hspace*{-0.5em}
    \subfloat[Unknown variance $\rho^2$]{
        \includegraphics[width=0.33\linewidth]{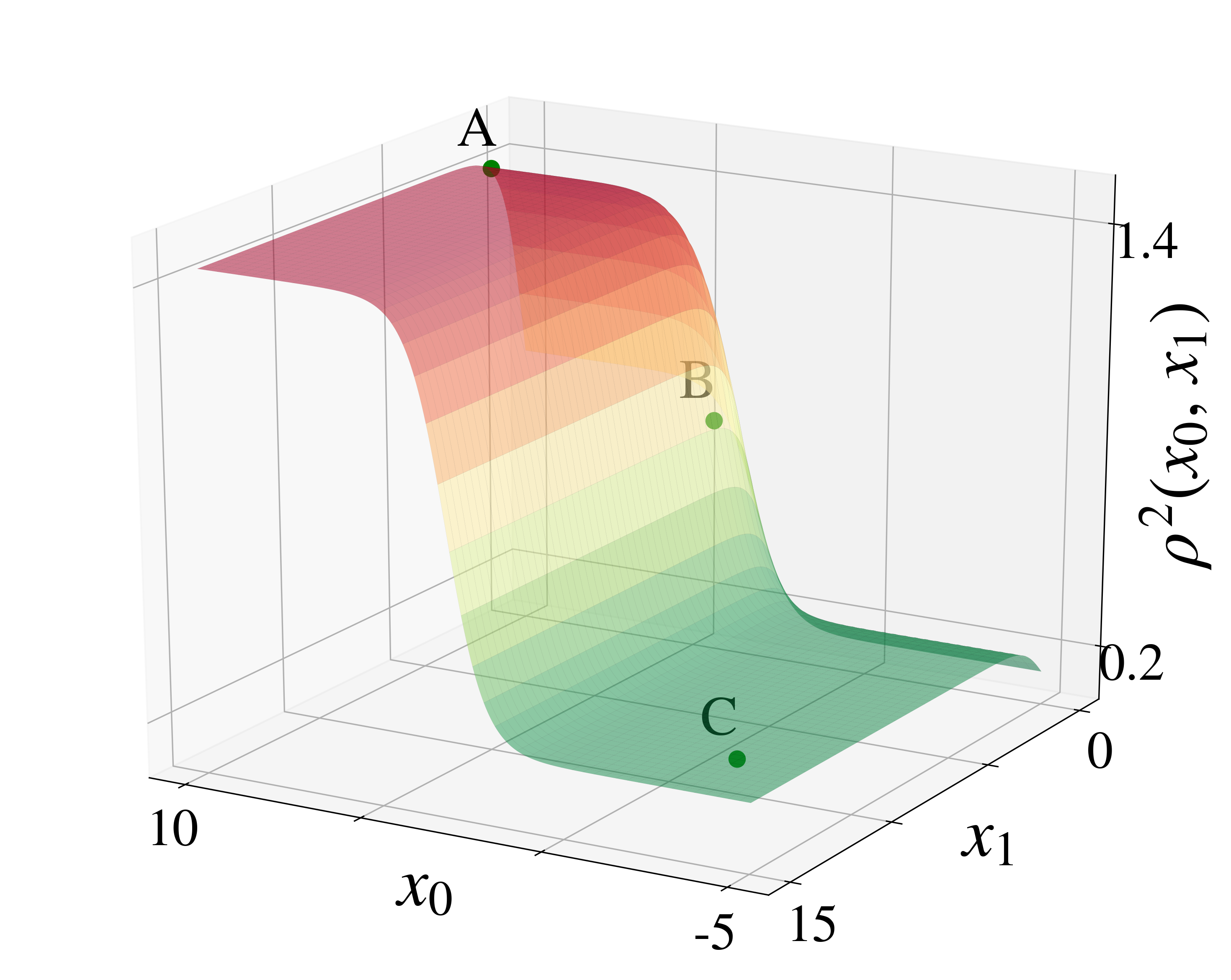}
        \label{fig:branin_var}
        }
    \subfloat[Histogram of variance]{
        \includegraphics[width=0.33\linewidth]{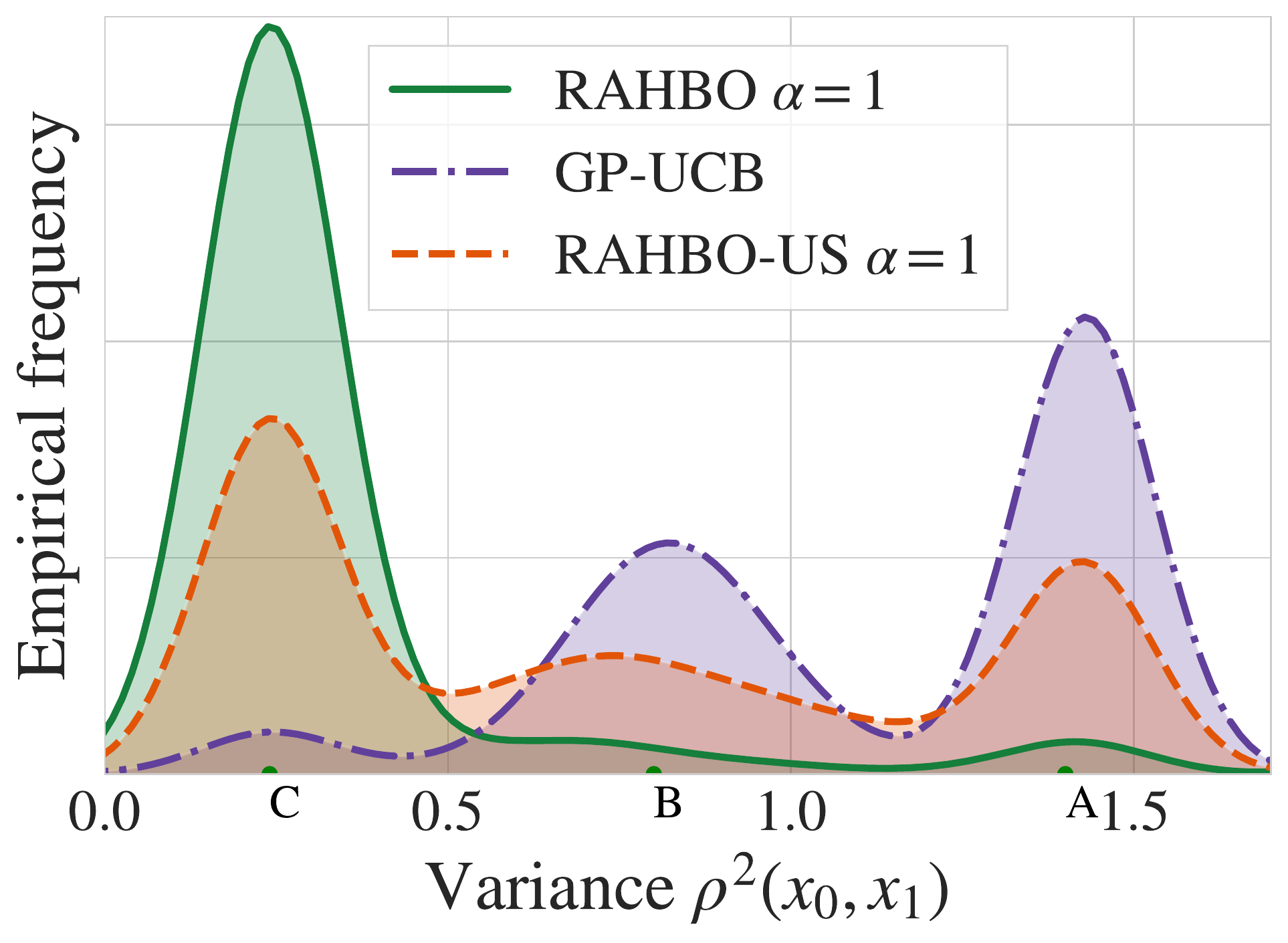}
        \label{fig:first_figure}
        }
     \caption{When there is a choice between identical optima with different noise level, standard BO tends to query points corresponding to higher noise. (a) Unknown objective with 3 global maxima marked as (A, B, C); (b) Heteroscedastic noise variance over the same domain: the noise level at (A, B, C) varies according to the sigmoid function; (c) Empirical variance distribution at all points acquired during BO procedure (over 9 experiments with different seeds). The three bumps correspond to the three global optima with different noise variance. \raucb dominates in choosing the risk-averse optimum, consequently yielding lower risk-averse regret in \Cref{fig:branin_cumregret}. \looseness -1}
     \label{fig:fig1}
     \vspace{0.2cm}
\end{figure}

\vspace{0.9cm}
\paragraph{Related work.}
\label{sec:background}
Bayesian optimization (BO) \cite{movckus1975} refers to approaches for optimizing a noisy black-box objective that is typically expensive to evaluate. A great number of BO methods have been developed over the years, including a significant number of variants of popular algorithms such as GP-UCB \cite{srinivas10}, Expected Improvement \cite{movckus1975}, 
and Thompson Sampling \cite{chowdhury2017kernelized}.
While the focus of standard BO approaches is mainly on trading-off exploration vs.~exploitation and optimizing for the expected performance, in this work, we additionally focus on the risk that is involved when working with noisy objectives, as illustrated in \Cref{fig:fig1}.


The vast majority of previous BO works assume (sub-) Gaussian and {\em homoscedastic} noise (i.e., input independent and of some known fixed level). Both assumptions can be restrictive in practice. For example, 
as demonstrated in \cite{cowenrivers2021empirical}, the majority of hyperparameter tuning tasks exhibit heteroscedasticity. A few works relax the first assumption and consider, e.g., heavy-tailed noise models \cite{heavytailedBO} and adversarially corrupted observations \cite{bogunovic2020corruption}. 
The second assumption is typically generalized via {\em heteroscedastic}  Gaussian process (GP), allowing an explicit dependence of the noise distribution on the evaluation points \cite{binois2016, bogunovic2016truncated,Binois2019ReplicationOE,kirschner2018information}. 
Similarly, in this work, we consider heteroscedastic GP models, but unlike the previous works, we specifically focus on the risk that is associated with querying and reporting noisy points. 



Several works have recently considered robust and risk-averse aspects in BO. Their central focus is on designing robust strategies and protecting against the change/shift in uncontrollable covariates. They study various notions including worst-case robustness \cite{bogunovic2018}, distributional robustness \cite{kirschner2020,nguyen2020distributionally}, robust mixed strategies \cite{sessa2020mixed} and other notions of risk-aversion~\cite{iwazaki2020meanvariance,riskbo_cakmak2020,nguyen2021valueatrisk}, and while some of them report robust regret guarantees, their focus is primarily on the robustness in the homoscedastic GP setting. Instead, in our setting, we account for the risk that comes from the realization of random noise with {\em unknown} distribution. Rather than optimizing the expected performance, in our risk-averse setting, we prefer inputs with {\em lower variance}. To this end, we incorporate the learning of the noise distribution into the optimization procedure via a \emph{mean-variance} objective. The closest to our setting is risk-aversion with respect to noise in multi-armed bandits \cite{Sani2012RiskAversionIM}. Their approach, however, fails to exploit correlation in rewards among similar arms.

\paragraph{Contributions.}  \looseness -1 We propose a novel \textit{Risk-averse Heteroscedastic Bayesian optimization} (\raucb) approach based on the optimistic principle that trades off the expectation and uncertainty of the mean-variance objective function.  We model both the objective and variance as (unknown) functions belonging to RKHS space of functions, and propose a practical risk-aware algorithm in the heteroscedastic GP setting. In our theoretical analysis, we establish rigorous sublinear regret guarantees for our algorithm, and provide a robust reporting rule for applications where only a single solution is required. We demonstrate the effectiveness of \raucb on synthetic benchmarks, as well as on hyperparameter tuning tasks for the Swiss free-electron laser and a machine learning model.



\section{Problem setting}
\label{sec:problem}
Let $\X$ be a given compact set of inputs ($\X\subset\R^d$ for some $d \in \mathbb{N}$).
We consider a problem of sequentially interacting with a fixed and unknown objective $f: \X \rightarrow \R$. 
At every round of this procedure, the learner selects an action $x_t \in \X$, and obtains a noisy observation 
\begin{equation} \label{eq:observational_model}
    y_t = f(x_t) + \xi(x_t),
\end{equation}
where $\xi(x_t)$ is zero-mean noise independent across different time steps $t$. 
In this work, we consider sub-Gaussian  
heteroscedastic noise that depends on the query location. 
\begin{definition}
\label{def:subG}
A zero-mean real-valued random variable $\xi$ is $\rho$--sub-Gaussian, if there exists variance-proxy $\rho^2$ such that
    $\forall \lambda \in \mathbb{R}, \ \  \E[e^{\lambda \xi}] \leq e^{\frac{\lambda^2 \rho^2}{2}}$. 
\end{definition}
For a sub-Gaussian $\xi$, its variance $\mathbb Var[\xi]$ lower bounds any valid variance-proxy $\rho$, i.e., $\mathbb Var[\xi]\leq \rho^2$. In case $\mathbb Var[\xi] = \rho^2$ holds,  $\xi$ is said to be \emph{strictly $\rho$--sub-Gaussian}. Besides zero-mean Gaussian random variables, most  standard symmetric bounded random variables (e.g., Bernoulli, beta, uniform, binomial) are strictly sub-Gaussian (see \cite[Proposition 1.1]{Arbel2019OnSS}).
Throughout the paper, we consider sub-Gaussian noise, and in \cref{sec:rahbo_unknown_var_prox}, we specialize to the case of strictly sub-Gaussian noise. 
\looseness=-1




\textbf{Optimization objective.} Unlike the previous works that mostly focus on sequential optimization of $f$ in the homoscedastic noise case, in this work, we consider the trade-off between risk and return in the heteroscedastic case. While there exist a number of risk-averse objectives, we consider the simple and frequently used mean-variance objective (MV) \cite{Sani2012RiskAversionIM}. 
Here, the objective value at $x\in \X$ is a trade-off between the (mean) return $f(x)$ and the risk expressed by its variance-proxy $\rho^2(x)$:
\looseness=-1
\begin{align}
    \mv(x) = f(x) - \alpha \rho^2(x), \label{eq:objective}
\end{align}
where $\alpha \geq 0$ is a so-called {\em coefficient of absolute risk tolerance}. In this work, we assume $\alpha$ is fixed and known to the learner. In the case of $\alpha=0$,  maximizing $\mv(x)$ coincides with the standard BO objective. 

\textbf{Performance metrics.} We aim to construct a sequence of input evaluations $x_t$ that eventually maximizes the risk-averse objective $\mv(x)$. To assess this convergence, we consider two metrics. The first metric corresponds to the notion of cumulative regret similar to the one used in standard BO and bandits. Here, the learner's goal is to maximize its risk-averse cumulative reward over a time horizon $T$, or equivalently minimize its \emph{risk-averse cumulative regret}:
\begin{equation}\label{eq:risk_averse_cumulative_regret}
    R_T = \sum_{t=1}^T \Bigl[\mv(x^*) - \mv(x_t)\Bigr],
\end{equation}
where $x^* \in \arg\max_{x\in \X} \mv(x)$. A sublinear growth of $R_T$ with $T$ implies vanishing average regret $R_T /T \rightarrow 0$ as $T \rightarrow \infty$. Intuitively, this implies the existence of some $t$ such that $\mv(x_t)$ is arbitrarily close to the optimal value $\mv(x^*)$.

The second metric is used when the learner seeks to simultaneously minimize the number of expensive function evaluations $T$. Namely, for a given accuracy $\epsilon \geq 0$, we report a single "good" risk-averse point $\xreported \in \X$ after a total of $T$ rounds, that satisfies:
\begin{equation}
    \mv(\xreported) \geq \mv(x^*) - \epsilon. 
    \label{eq:simple_regret}
\end{equation}

Both metrics are important for choosing risk-averse solutions and which one is preferred depends on the application at hand. For example, risk-averse cumulative regret $R_T$ might be of a greater interest in online recommendation systems, while reporting a single point with high MV value might be more suitable when tuning machine learning hyperparameters. We consider both performance metrics in our experiments.\looseness=-1


\textbf{Regularity assumptions.} We consider standard smoothness assumptions \cite{srinivas10,bogunovic2018} when it comes to the unknown function $f:\X \rightarrow \R$. In particular, we assume that $f(\cdot)$ belongs to a reproducing kernel Hilbert space (RKHS) $\mathcal{H}_\kappa$ (a space of smooth and real-valued functions defined on $\X$), i.e., $f \in \mathcal{H}_\kappa$, induced by a kernel function $\kappa(\cdot, \cdot)$. We also assume that $\kappa(x,x') \leq 1$ for every $x,x' \in \X$. Moreover, the RKHS norm of $f(\cdot)$ is assumed to be bounded $\|f\|_{\kappa} \leq \mathcal{B}_f$ for some fixed constant $\mathcal{B}_f>0$. 
We assume that the noise $\xi(x)$ is $\rho(x)$--sub-Gaussian with variance-proxy  $\rho^2(x)$ uniformly bounded $\rho(x) \in [\rhomin, \rhomax]$ for some constant values $\rhomax \geq\rhomin > 0$.

\section{Algorithms}
\label{sec:main}

We first recall the Gaussian process (GP) based framework for sequential learning of RKHS functions from observations with heteroscedastic noise. Then, in \Cref{sec:warm_up}, we consider a simple risk-averse Bayesian optimization problem with \emph{known} variance-proxy, and later on in \Cref{sec:rahbo_unknown_var_prox}, we focus on our main problem setting in which the variance-proxy is \emph{unknown}.

\subsection{Bayesian optimization with heteroscedastic noise}
Before addressing the risk-averse objective, we briefly recall the standard GP-UCB algorithm \citep{srinivas10} in the setting of heteroscedastic sub-Gaussian noise.  The regularity assumptions  permit the construction of confidence bounds via GP model. Particularly, to decide which point to query at every round, GP-UCB makes use of the posterior GP mean and variance denoted by $\mu_t(\cdot)$ and $\sigma^2_t(\cdot)$, respectively. They are computed based on the previous measurements $y_{1:t} = [y_1, \dots, y_t]^\top$ and the given kernel $\kappa(\cdot,\cdot)$ :\looseness=-1 
\begin{align}
    &\mu_t(x) = \kappa_t(x)^T(K_t + \lambda \Si_t)^{-1} y_{1:t} \label{eq:gp_mean_hetero},\\
    & \sigma^2_t(x) = \frac{1}{\lambda} \big(\kappa(x,x) - \kappa_t(x)^\top(K_t + \lambda \Si_t)^{-1}\kappa_t(x) \big) \label{eq:gp_var_hetero},
\end{align} 
where $\Si_t := \text{diag}(\rho^2(x_1), \dots, \rho^2(x_t))$, $(K_t)_{i,j} = \kappa(x_i, x_j), \ \kappa_t(x)^T = [\kappa(x_1, x), \dots, \kappa(x_t, x) ]^T$, $\lambda > 0$ and
prior modelling assumptions are $\xi(\cdot) \sim \mathcal{N}(0,\rho^2(\cdot))$ and $f \sim GP(0, \lambda^{-1}\kappa)$.

At time $t$, GP-UCB maximizes the upper confidence bound of $f(\cdot)$, i.e., \looseness=-1 
\begin{align}
    x_t \in \argmax_{x\in \X}\;\underbrace{\mu_{t-1}(x) + \beta_t \sigma_{t-1}(x)}_{=:  \mathrm{ucb}^f_t(x)}.
    \label{eq:ucb_acq}
\end{align} \looseness=-1 
If the noise $\xi_t(x_t)$ is heteroscedastic  and $\rho(x_t)$-sub-Gaussian, the following confidence bounds hold:  
\begin{lemma}[Lemma 7 in \cite{kirschner2018information}] \label{lemma:kirschner}
Let $f\in \mathcal{H}_\kappa$, and $\mu_t(\cdot)$ and $\sigma^2_t(\cdot)$  be defined as in \cref{eq:gp_mean_hetero,eq:gp_var_hetero} with $\lambda >0.$ Assume that the observations $(x_t, y_t)_{t \geq 1}$  satisfy \cref{eq:observational_model}. Then the following holds for all $t \geq 1$ and $x \in \X$ with probability at least $1-\delta$:
\begin{align}\label{def:beta}
    &|\mu_{t-1}(x) - f(x)| \leq 
    \underbrace{\Bigg( 
        \sqrt{2\ln \bigg( \frac{\det(\lambda \Si_t +K_t)^{1/2}}{\delta\det(\lambda \Si_t)^{1/2}}}
        \bigg)
        + \sqrt{\lambda}\|f\|_{\kappa}
    \Bigg)}_{:=\beta_{t}} \sigma_{t-1}(x). 
\end{align} 
\end{lemma}
Here, $\beta_t$ stands for the  parameter that balances between exploration vs.~exploitation and ensures the validity of confidence bounds. The analogous concentration inequalities in case of homoscedastic noise were considered in \cite{abbasi2012,chowdhury2017kernelized,srinivas10}.

\textbf{Failure of GP-UCB in the risk-averse setting.} GP-UCB is guaranteed to achieve sublinear cumulative regret with high probability in the risk-neutral (homoscedastic/heteroscedastic) BO setting \cite{srinivas10,chowdhury2017kernelized}. However, for the risk-averse setting in \cref{eq:objective}, the maximizers $x^* \in \argmax_{x \in \X} \mv(x)$ and $x^*_{f} \in \argmax_{x \in \X} f(x)$ might not coincide, and consequently, $\mv(x^*)$ can be significantly larger than $\mv(x^*_{f})$. This is illustrated in \Cref{fig:fig1}, where GP-UCB most frequently chooses optimum $A$ of the highest risk.




\subsection{Warm up: Known variance-proxy} 
\label{sec:warm_up}
We remedy the previous issue with GP-UCB by proposing a natural \emph{Risk-averse Heteroscedastic} BO (\raucb) in case of the known variance-proxy $\rho^2(\cdot)$. At each round $t$, \raucb chooses the action:  
\begin{equation} \label{eq:ra_gp_ucb_first}
    x_t \in \argmax_{x\in \X}\; \mu_{t-1}(x) + \beta_t \sigma_{t-1}(x) - \alpha \rho^2(x),
\end{equation}
where $\beta_t$ is from \cref{lemma:kirschner} and $\alpha$ is from \cref{eq:objective}. In the next section, we further relax the assumption of the variance-proxy and consider a more practical setting when $\rho^2(\cdot)$ 
is unknown to the learner. For the current setting, the performance of \raucb is formally captured in the following proposition. 

\begin{prop} \label{proposition_known}
   Consider any $f\in \mathcal{H}_{\kappa}$ with $\|f\|_{\kappa} \leq \mathcal{B}_f$ and sampling model from \cref{eq:observational_model} with known variance-proxy $\rho^2(x)$.  Let $\lbrace \beta_t\rbrace_{t=1}^T$ be set as in \cref{lemma:kirschner} with $\lambda=1$. 
Then, with probability at least $1-\delta$, \raucb attains cumulative risk-averse regret $R_T = \mathcal{O}\big(\beta_T \sqrt{T  \gamma_T(\rhomax^2 +1) }\big)$.
\end{prop} 
\looseness=-1

Here, $\gamma_T$ denotes the \emph{maximum information gain} \cite{srinivas10} at time $T$ defined via mutual information $I(y_{1:T}, f_{1:T})$ between evaluations $y_{1:T}$ and $f_{1:T}= [f(x_1), \dots, f(x_T)]^\top $ at points $A \subset D$: \looseness=-1
\begin{align}
& \gamma_T := \underset{A\subset\mathcal{X},\; |A| = T}{\max} \; I(y_{1:T}, f_{1:T}), \label{eq:gain}\\
& \text{where}\ \  I(y_{1:T}, f_{1:T}) = \frac{1}{2}\sum_{t=1}^T \ln\Big( 1+ \tfrac{\sigma_{t-1}^2(x_t)}{   \rho^2(x_t)}\Big) \label{eq:mutual_info}
\end{align} 
 in case of heteroscedastic noise (see \cref{sec:proof_known_rho}). The upper bounds on $\gamma_T$ are provided in \cite{srinivas10} widely used kernels. These upper bounds typically scale sublinearly in $T$; for linear kernel $\gamma_T = \mathcal O(d\log T)$, and in case of squared exponential kernel $\gamma_T = \mathcal O(d(\log T)^{d+1})$. While these bounds are derived assuming the homoscedastic GP setting with some fixed constant noise variance, we show (in \cref{sec:gamma_bound}) that the same rates (up to a multiplicative constant factor) apply in the heteroscedastic case. \looseness-1 


\subsection{RAHBO for unknown variance-proxy}
\label{sec:rahbo_unknown_var_prox}
In the case of unknown variance-proxy, the confidence bounds for the unknown $f(x)$ in \cref{lemma:kirschner} can not be readily used, and we construct new ones on the combined mean-variance objective. To learn about the unknown $\rho^2(x)$, we make some further assumptions.
\begin{assumption}
    The variance-proxy $\rho^2(x)$ belongs to an RKHS induced by some kernel $\kappa^{\varproxy}$, i.e., $\rho^2 \in \mathcal{H}_{\kappa^{\varproxy}}$, and  its RKHS norm is bounded $ \| \rho^2\|_{\kappa^{\varproxy}} \leq \mathcal{B}_{\varproxy}$ for some finite $\mathcal{B}_{\varproxy}>0$. Moreover, the noise $\xi(x)$ in \cref{eq:observational_model} is strictly $\rho(x)$--sub-Gaussian, i.e., $\mathbb Var[\xi(x)] = \rho^2(x)$ for every $ x \in \X$.
    \label{asm:ssubG}
\end{assumption}
As a consequence of our previous assumption, we can now focus on estimating the variance since $\mathbb{V}ar[\xi(\cdot)]$ and $\rho^2(\cdot)$ coincide. In particular, to estimate $\mathbb Var[\xi(\cdot)]$ we consider a \emph{repeated experiment setting}, where for each $x_t$ we collect $k>1$ evaluations $\{y_i(x_t)\}_{i=1}^{\krepeat}$, $y_i(x_t) = f(x_t) + \xi_i(x_t)$. Then, the sample mean and variance of $\xi(x_t)$ are given as: 
\begin{equation} \label{eq:emp_mean_and_var}
    \hat{m}_{\krepeat}(x_t)= \frac{1}{\krepeat} \sum_{i=1}^{\krepeat} y_i(x_t) \quad \text{and} \quad
    \hat{s}^2_{\krepeat}(x_t)= \frac{1}{\krepeat - 1} \sum_{i=1}^{\krepeat}\big(y_i (x_t) - \hat{m}_{\krepeat}(x_t)\big)^2.
\end{equation}



The key idea is that for strictly sub-Gaussian noise $\xi(x)$, $\hat{s}^2_{1:t} = [\hat{s}^2_{\krepeat}(x_1), \dots, \hat{s}^2_{\krepeat}(x_t)]^\top$ yields {\em unbiased, but noisy} evaluations of the unknown variance-proxy $\rho^2_{1:t} = [\rho^2(x_1), \dots, \rho^2(x_t)]^\top$, i.e., \looseness -1 
\begin{equation}
    \hat{s}^2_{\krepeat}(x_t) = \rho^2(x_t) + \eta(x_t)
    \label{eq:sample_variance_eval}
\end{equation}    
with zero-mean noise $\eta(x_t)$. In order to efficiently estimate $\rho^2(\cdot)$, we need an additional assumption.\looseness=-1
\begin{assumption} \looseness -1 
The noise $\eta(x)$ in \cref{eq:sample_variance_eval} is  $\rho_{\eta}(x)$--sub-Gaussian with known $\rho_{\eta}^2(x)$ and the realizations $\{\eta(x_t)\}_{t\geq 1}$ are independent between $t$. \label{asm:subG_known}
\end{assumption}
We note that a similar assumption is made in \cite{Sani2012RiskAversionIM} in the multi-armed bandit setting. The fact that $\rho^2_{\eta}(\cdot)$ is known is rather mild as Assumption~\ref{asm:ssubG} allows controlling its value.
For example, in case of strictly sub-Gaussian $\eta(x)$ we show (in \cref{sec:varproxy_for_var}) that $\mathbb Var[ \eta(\cdot)] =  \rho^2_{\eta}(\cdot)  \leq  2\rho^4(\cdot)/(k-1)$. Then, given that $\rho^2(\cdot)\leq \rhomax^2$, we can utilize the following (rather conservative) bound as a variance-proxy, i.e.,  $\rho_{\eta}^2(x)=2\rhomax^4/(k-1)$.  

 \begin{algorithm}[t]
    \caption{Risk-averse Heteroscedastic Bayesian Optimization (\raucb)}
    \label{alg:risk-averse-bo}
    \begin{algorithmic}[1]  
                \REQUIRE Parameters $\alpha, \{\beta_t, \beta_t^{\varproxy}\}_{t\geq1}, \lambda$, $k$, Prior $\mu_0^f = \mu_0^{\varproxy} =0$, \text{Kernel functions} $\kappa, \kappa^{\varproxy}$
            	\FOR{$t = 1, 2, \hdots$}
            	   \STATE Construct confidence bounds $\mathrm{ucb}_t^{\varproxy}(\cdot)$ and $\mathrm{lcb}_t^{\varproxy}(\cdot)$ as in  \cref{eq:ucb_rho,eq:lcb_rho}
            	    \STATE Construct $\mathrm{ucb}^f_t(\cdot)$ as in \cref{eq:ucb_acq} 
                    \STATE Select 
                       $ x_t \in \argmax_{x \in \X} \mathrm{ucb}^f_{t}(x) - \alpha \;\mathrm{lcb}_{t}^{\varproxy}(x)$
                    \STATE Observe $k$ samples:
                       $ y_i(x_t) = f(x_t) + \xi_i(x_t)\quad \text{for every} \quad i \in [k]$
                    \STATE Use samples $\{y_i(x_t)\}_{i=1}^k$ to compute sample mean $\hat{m}_{\krepeat}(x_t)$ and variance $\hat{s}^2_{\krepeat}(x_t)$ as in 
                    \cref{eq:emp_mean_and_var}
                    \STATE Use  $x_t, \hat s_k^2(x_t)$ to update posterior $\mu_t^{var}(\cdot)$ and  
                    $\sigma_t^{var}(\cdot)$ as in \cref{eq:gp_mean_hetero_1,eq:gp_var_hetero_1}  
                    \STATE Use $\mathrm{ucb}_t^{\varproxy}(\cdot)$ to compute $\hat \Si_t$ as in \cref{eq:hatsigma}
                    \STATE Use $x_t, \hat m_k(x_t)$ and $\hat \Si_t$ to update posterior $\mu_t(\cdot)$ and $\sigma_t(\cdot)$ as in \cref{eq:gp_mean_hetero,eq:gp_var_hetero}  
                    
                \ENDFOR 
            \end{algorithmic}
      \end{algorithm}

\textbf{\raucb algorithm.} We present our Risk-averse Heteroscedastic BO approach for unknown variance-proxy in \cref{alg:risk-averse-bo}. Our method relies on building the following two GP models. 

Firstly, we use sample variance evaluations  $\hat{s}^2_{1:t}$ to construct a GP model for $\rho^2(\cdot)$. The corresponding  $\mu^{\varproxy}_{t-1}(\cdot)$ and  $\sigma^{\varproxy}_{t-1}(\cdot)$ are computed as in \cref{eq:gp_mean_hetero,eq:gp_var_hetero} by using kernel $\kappa^{\varproxy}$, variance-proxy $\rho^2_{\eta}(\cdot)$ and noisy observations $\hat{s}^2_{1:t}$. Consequently, we build the upper and lower confidence bounds $\mathrm{ucb}^{var}_t(\cdot)$ and $\lcb^{var}_t(\cdot)$ of the variance-proxy $\rho^2(\cdot)$ and we set $\beta_t^{\varproxy}$ according to~\cref{lemma:kirschner}:
\begin{align}
    \mathrm{ucb}_t^{\varproxy}(x) &:= \mu^{\varproxy}_{t-1}(x) + \beta_t^{\varproxy} \sigma^{\varproxy}_{t-1}(x) \label{eq:ucb_rho},\\
    \mathrm{lcb}_t^{\varproxy}(x) &:= \mu^{\varproxy}_{t-1}(x) - \beta_t^{\varproxy} \sigma^{\varproxy}_{t-1}(x)\label{eq:lcb_rho}.
\end{align}

 Secondly, we use sample mean evaluations $\hat{m}_{1:t} = [\hat{m}_{\krepeat}(x_1),\dots, \hat{m}_{\krepeat}(x_t)]^\top$ to construct a GP model for $f(\cdot)$. 
 The mean $\mu_t(\cdot)$ and variance $\sigma_t^2(\cdot)$ in \cref{eq:gp_var_hetero,eq:gp_mean_hetero}, however, rely on the unknown variance-proxy $\rho^2(\cdot)$ in $\Si_t$, an we thus use its upper confidence bound $\mathrm{ucb}^{\varproxy}_t(\cdot)$ truncated with $\rhomax^2$:
 \begin{align}\label{eq:hatsigma}
     \hat \Si_t := \tfrac{1}{k}\text{diag}\big(\min\{\mathrm{ucb}^{\varproxy}_t(x_1), \rhomax^2\} ,\ldots, \min\{\mathrm{ucb}^{\varproxy}_t(x_t), \rhomax^2\}\big),
 \end{align}
where $\hat \Si_t$ is corrected by $k$ since every evaluation in $\hat{m}_{1:t}$ is an average over $k$ samples. This substitution of the unknown variance-proxy by its conservative estimate guarantees that the confidence bounds $\mathrm{ucb}^f_t(x):= \mu_{t-1}(x) + \beta_t \sigma_{t-1}(x)$ on $f$ also hold with high probability (conditioning on the confidence bounds for $\rho(\cdot)$ holding true; see \cref{app:method} for more details). 
 
Finally, we define the acquisition function as $\mathrm{ucb}^{\mv}_t(x) := \mathrm{ucb}^f_t(x) - \alpha\mathrm{lcb}_t^{\varproxy}(x)$, i.e., selecting $x_t \in \arg \max_{x \in \X} \mathrm{ucb}^{\mv}_t(x)$ at each round $t$. 

The proposed algorithm leads to new maximum information gains $\hat\gamma_T = \max_{A}  I(\hat m_{1:T}, f_{1:T})$ and $\Gamma_T = \max_{A}  I(\hat s^2_{1:T}, \rho^2_{1:T})$ for sample mean $\hat m_{1:T}$ and sample variance $\hat s^2_{1:T}$ evaluations. The corresponding mutual information in $\hat\gamma_T$ and $\Gamma_T$ is computed according to \cref{eq:mutual_info} for heteroscedastic noise with variance-proxy $\rhomax^2/k$ and $\rho^2_{\eta}$, respectively (see \cref{app:proof_unknown_rho}). 
The performance of \raucb is captured in the following theorem. 

\begin{customthm}{1}\label{theorem}\textit{
Consider any $f\in \mathcal{H}_{\kappa}$ with $\|f\|_{\kappa} \leq \mathcal{B}_f$ and sampling model in \cref{eq:observational_model} with unknown variance-proxy $\rho^2(x)$ that satisfies Assumptions~\ref{asm:ssubG} and \ref{asm:subG_known}. Let $\{x_t\}_{t=1}^T$ denote the set of actions chosen by \raucb (\cref{alg:risk-averse-bo}) over T rounds.
 Set  $\lbrace \beta_t\rbrace_{t=1}^T$ and $\lbrace \beta^{var}_t\rbrace_{t=1}^T$ according to \cref{lemma:kirschner} with $\lambda=1$, $\mathcal{R}^2 = \max_{x\in \mathcal X} \rho^2_{\eta}(x_t)$ and $\rho(\cdot) \in [\rhomin, \rhomax]$.  Then, the risk-averse cumulative regret $R_{T}$ of \raucb is bounded as follows}: 
\begin{align}
& \mathrm{Pr} 
\bigg\{
R_{T}  \leq 
 \beta_T k\sqrt{\frac{2T\hat\gamma_T}{ \ln (1 + k/\rhomax^{2})}} + \alpha \beta_T^{var} k\sqrt{\frac{2T\Gamma_T }{\ln (1 + \mathcal{R}^{-2})}} 
,\quad \forall T \geq 1 \bigg\} \geq 1 - \delta.
\end{align}
\end{customthm}

The risk-averse cumulative regret of \raucb depends sublinearly on $T$ for most of the popularly used kernels. This follows from 
the implicit sublinear dependence on $T$ in $\beta_T, \beta_T^{\varproxy}$ and $\hat\gamma_T,\Gamma_T$ (the bounds in case of heteroscedastic noise replicate the ones used in Proposition~\ref{proposition_known} as shown in \Cref{sec:beta_bound,sec:gamma_bound}). Finally, the result of Theorem~\ref{theorem} provides a non-trivial trade-off for number of repetitions $k$ where larger $k$ increases sample complexity but also leads to better estimation of the noise model. Furthermore, we obtain the bound for the number of rounds $T$ required for identifying an $\epsilon$-optimal point:\looseness=-1


\begin{corollary}
Consider the setup of Theorem~\ref{theorem}. Let $A = \{x_t\}_{t=1}^T$ denote actions selected by \raucb over $T$ rounds. Then, with probability at least $1-\delta$, the reported point $\xreported: =  \arg\max_{x_t \in A} \lcb^{\mv}_t(x_t),$ where $ \lcb^{\mv}_t(x_t) = \mathrm{lcb}^f_{t}(x) - \alpha \;\mathrm{ucb}_{t}^{\varproxy}(x)$,  achieves $\epsilon$-accuracy, i.e., $\mv(x^*) - \mv(\xreported)\leq \epsilon$, after $T \geq\tfrac{ 4\beta_T^2\hat\gamma_T/\ln (1+   k/\rhomax^{2}) + 4\alpha (\beta_t^{var})^2\Gamma_T/\ln (1+   \mathcal{R}^{-2}) }{\epsilon^2}$ rounds. 
  \label{col_report}
\end{corollary}

The previous result demonstrates the sample complexity rates when a single risk-averse reported solution is required.  
We note that both Theorem~\ref{theorem} and \cref{col_report} provide guarantees for choosing risk-averse solutions, and depending on application at hand, we might consider either one of the proposed performance metrics. We demonstrate use-cases for both in the following section.


\section{Experiments}
\label{sec:experiments}

In this section, we experimentally validate \raucb on two synthetic examples and two real hyperparameter tuning tasks, and compare it with the baselines. We provide an open-source implementation of our method.\footnote{\url{https://github.com/Avidereta/risk-averse-hetero-bo}} \looseness-1

\textbf{Baselines.} We compare against two baselines: 
As the first baseline, we use \gpucb with heteroscedastic noise as a standard risk-neutral algorithm that optimizes the unknown $f(x)$. As the second one, we consider a risk-averse baseline that uniformly learns variance-proxy $\rho^2(x)$ \emph{before} the optimization procedure, in contrast to RAHBO which learns the variance-proxy on the fly.  We call it \raucbus, standing for \raucb with uncertainty sampling. It consists of two stages: (i) uniformly learning $\rho^2(x)$ via uncertainty sampling,  (ii) \gpucb applied to the mean-variance objective, in which instead of the unknown $\rho^2(x)$ we use the mean of the learned model. Note that \raucbus is the closest to the contextual BO setting \cite{iwazaki2020meanvariance}, where the context distribution is assumed to be known. \looseness-1

\textbf{Experimental setup.} At each iteration $t$, an algorithm queries a point $x_t$ and observes sample mean and sample variance of $k$ observations  $\{y_i(x_t)\}_{i=1}^k$. We use a heteroscedastic GP for modelling $f(x)$ and a homoscedastic GP for $\rho^2(x)$. We set $\lambda=1$ and $\beta_t = 2$, which is commonly used in practice to improve performance over the theoretical results. Before the BO procedure, we determine the GP hyperparameters maximizing the marginal likelihood. 
To this end, we use initial points that are same for all the baselines and are chosen via Sobol sequence that generates low discrepancy quasi-random samples. We repeat each experiment several times, generating new initial points for every repetition. We use two metrics: (a) risk-averse cumulative regret $R_t$ computed for the acquired inputs; (b) simple regret $\mv(x^*) - \mv(\xreported)$ computed for inputs as reported via \cref{col_report}.  For each metric, we report its mean $\pm$ two standard errors over the repetitions.

\begin{figure}[t]
    \subfloat[Illustration of the sine function (left) and noise variance (right)]{
        \includegraphics[width=\textwidth]{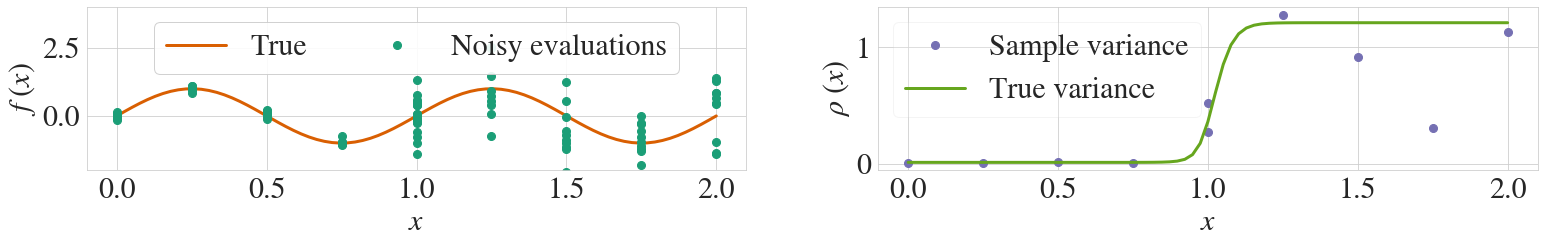}
    \label{fig:sine}
        }\\
    \subfloat[Cumulative regret]{
        \includegraphics[width=0.28\linewidth]{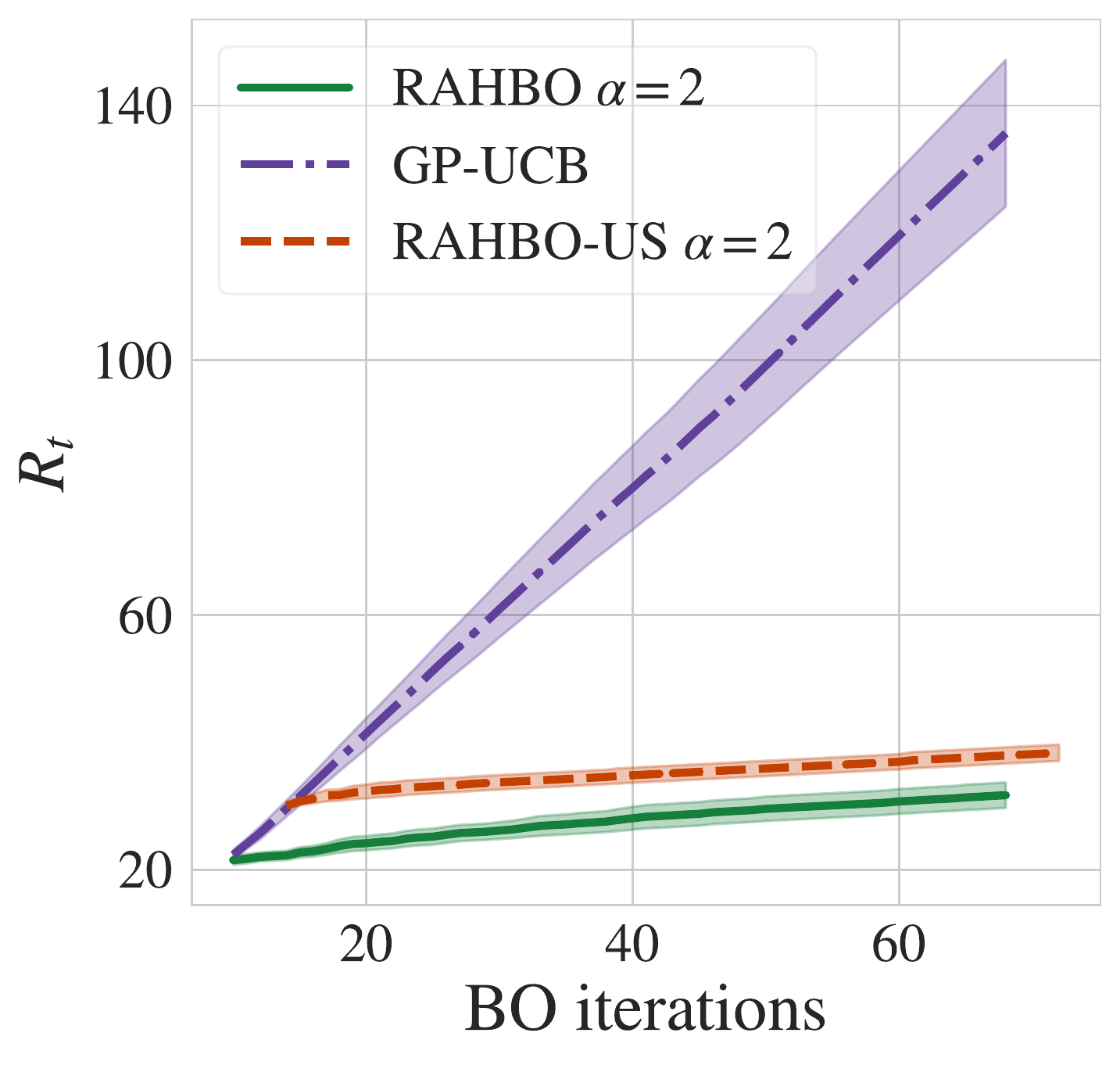}
        \label{fig:sin_cumregret}
        }
    \subfloat[Suboptimality w.r.t.~MV]{
        \includegraphics[width=0.28\linewidth]{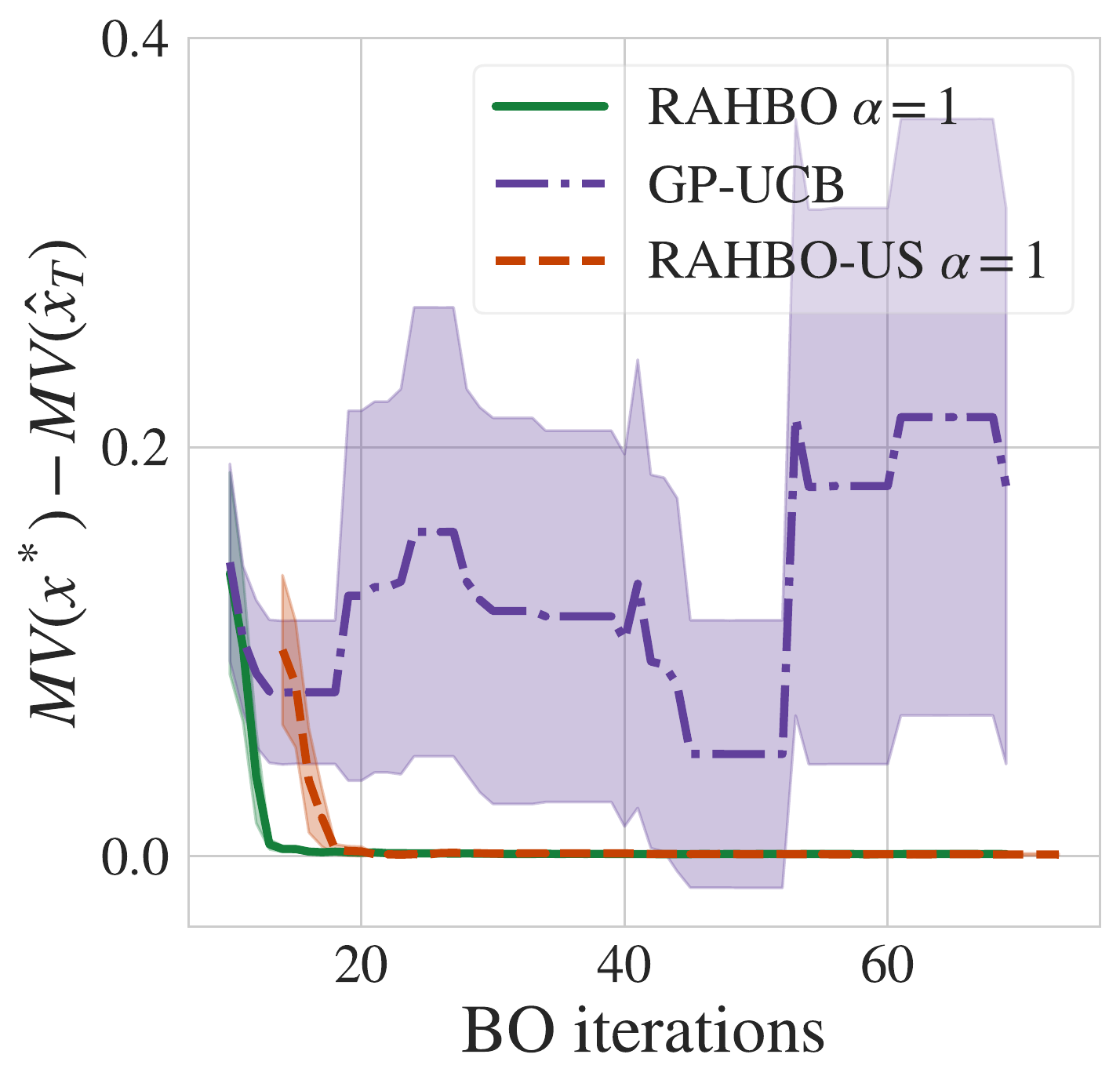}
        \label{fig:sine_mv}
        }
    \subfloat[Suboptimality w.r.t.~$f$]{
        \includegraphics[width=0.28\linewidth]{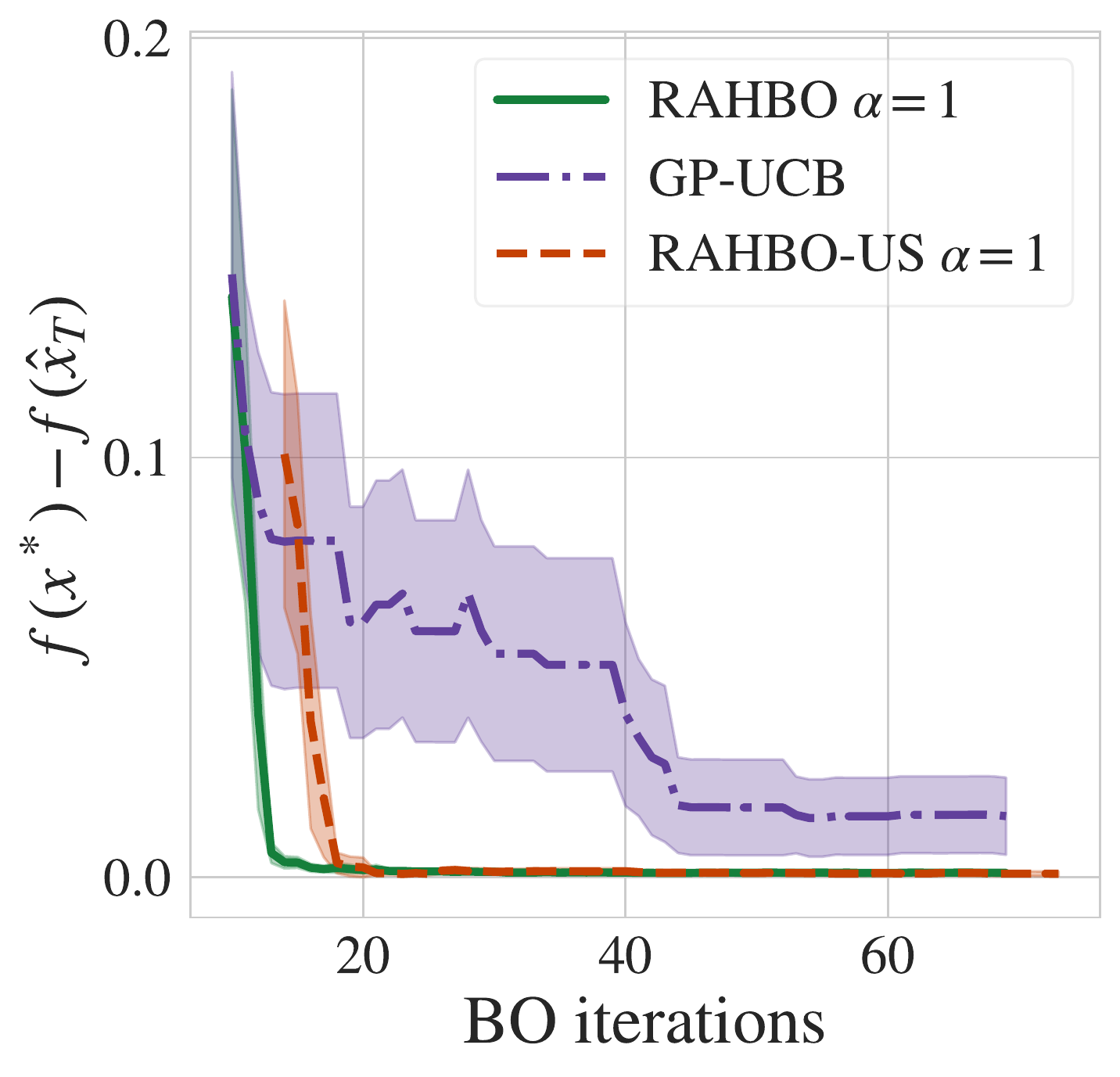}
        \label{fig:sine_f}
        }
    \caption{(a) Unknown true objective along with noisy evaluations with varying noise level (left) and unknown true noise variance and its evaluations (right).  (b) Cumulative regret. (c) Simple \mv regret for reporting rule $\xreported = \argmax_{x_t}\mathrm{lcb}_T(x_t)$.  (c)  Simple regret $f(x^*) - f(\xreported)$ for the unknown function at the reported point $\xreported$ from (d). \raucb not only leads to strong results in terms of \mv but also in terms of the mean objective $f(x)$. }
    \label{fig:sine_res}
\end{figure}

\textbf{Example function } We first illustrate the methods performance on a sine function depicted in \Cref{fig:sine}. This function has two global optimizers. We induce a heteroscedastic zero-mean Gaussian noise on the measurements. We use a sigmoid function for the noise variance, as depicted in \Cref{fig:sine}, that induces small noise on $[0,1]$ and higher noise on $(1,2]$. We initialize the algorithms by selecting $10$ inputs $x$ at random and keep these points the same for all the algorithms. We use $k = 10$ samples at each chosen $x_t$.  The number of acquisition rounds is $T = 60$. We repeat the experiment $30$ times for each method and show their average performances in  \Cref{fig:sine_res}. 

\begin{figure}[t]
    \vspace{-2mm}
    \includegraphics[width=1.\linewidth]{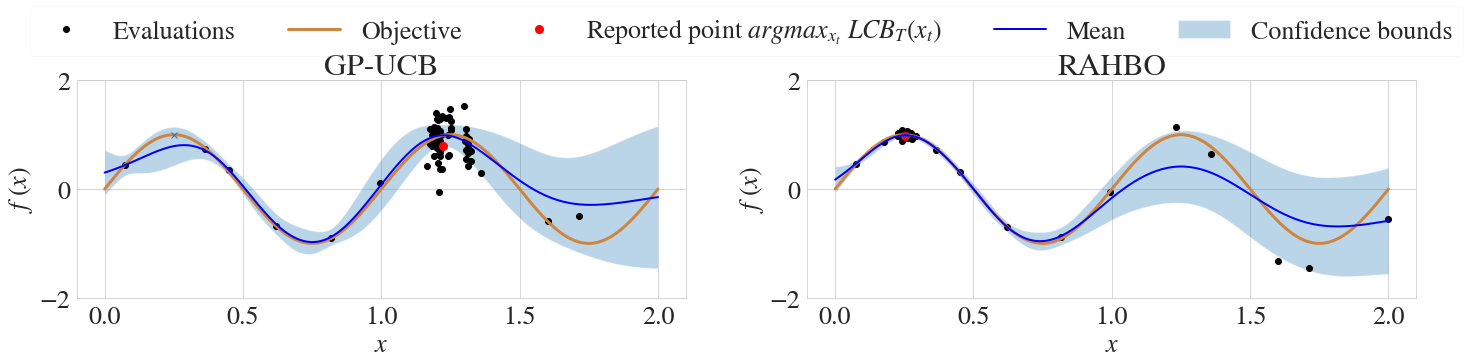}
        \label{fig:sine_gp1}\vspace{-5mm}
    \caption{GP models fitted for \gpucb  (left) and \raucb (right) for sine function. After  initialization with the same sampled points, \gpucb  concentrates on the high-noise region whereas \raucb prefers small variance. Additional plots are presented in \cref{app:exp_additional}. }
\end{figure}

\textbf{Branin benchmark\ \ } \looseness -1 Next, we evaluate the methods on the (negated) Branin benchmark function in \Cref{fig:branin}, achieving its optimum value $f^*= -0.4$ at $(-\pi, 12.3), (\pi, 2.3), (9.4, 2.5)$. The heteroscedastic variance function illustrated in \Cref{fig:branin_var} defines different noise variances for the three optima. 
We initialize all algorithms by selecting $10$ inputs. We use $k = 10$ samples to estimate the noise variance.  The number of acquisition rounds is $T = 150$. We repeat BO $25$ times and show the results in  \Cref{fig:branin_cumregret,fig:first_figure}. \Cref{fig:first_figure} provides more intuition behind the observed regret: 
UCB exploits the noisiest maxima the most, while \raucb prefers smaller variance. 
\textbf{Tuning Swiss free-electron laser\ \ } In this experiment, we tune the parameters of Swiss X-ray free-electron laser (\swissfel), an important scientific instrument that generates very short pulses of X-ray light and enables researchers to observe extremely fast processes.  The main objective is to maximize the pulse energy measured by a gas detector, that is a time-consuming and repetitive task during the SwissFEL operation. Such (re-)tuning takes place while user experiments on SwissFEL are running, and thus the cumulative regret is the metric of high importance in this application.


\looseness -1 We use real \swissfel measurements collected in \cite{kirschner19a} 
to train a neural network surrogate model, and use it to simulate the \swissfel objective 
$f(x)$ for new parameter settings $x$. We similarly fit a model of the heteroscedastic variance by regressing the squared residuals via a GP model. Here, we focus on the calibration of the four most sensitive parameters.

\begin{figure}[t]
    \subfloat[\looseness -1 Empirical distribution of true $f(x)$ (left) and $\rho^2(x)$ (right) for \swissfel]{
    \includegraphics[width=1.\linewidth]{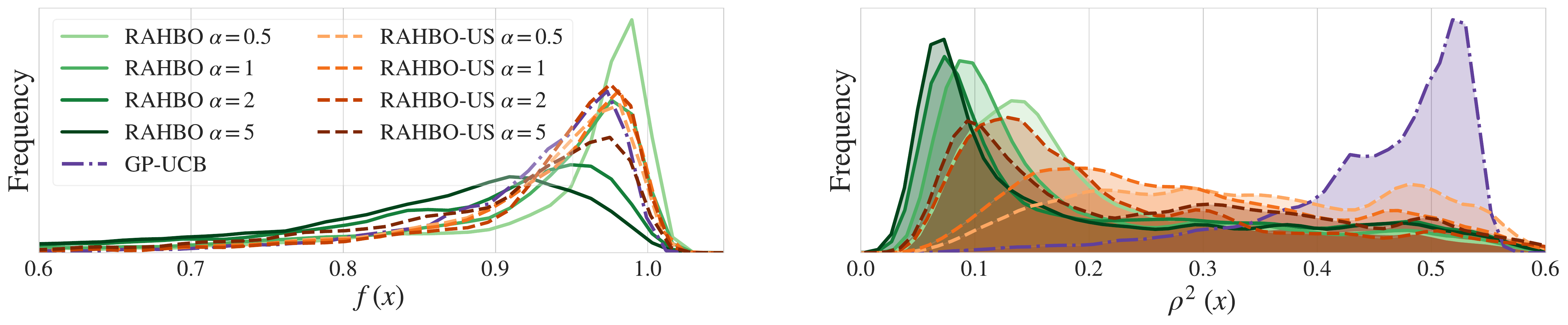}
       \label{fig:swissfel_hist}\vspace{-2mm}
    \label{fig:swissfel_hist}
    } \\
 \hspace{-1cm}
    \subfloat[Mean-variance tradeoff]{
    \includegraphics[width=0.48\linewidth]{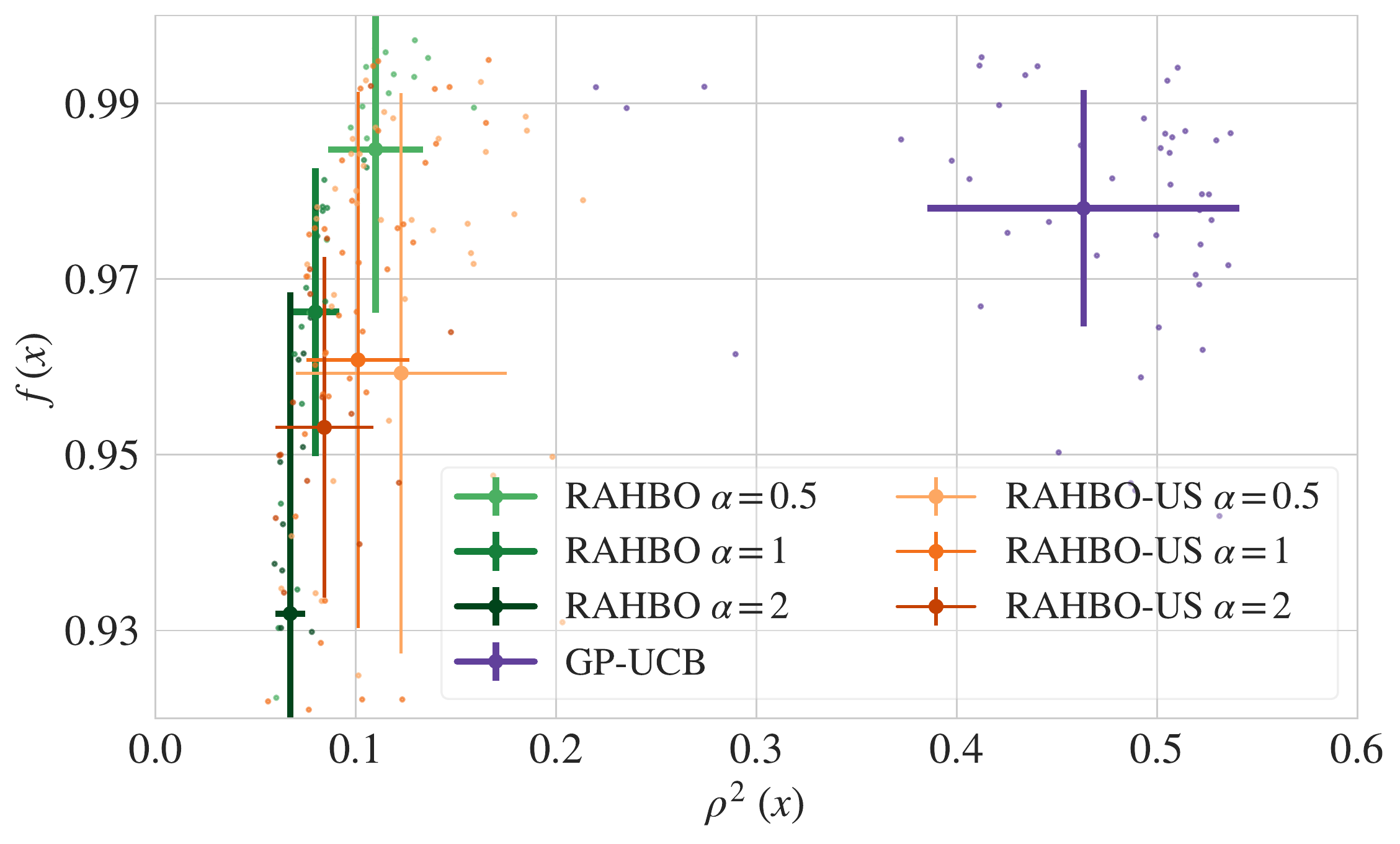}
    \label{fig:swissdel_reported}
    }
    \subfloat[Cum. regret ($\alpha=0.5$)]{
    \includegraphics[width=0.26\linewidth]{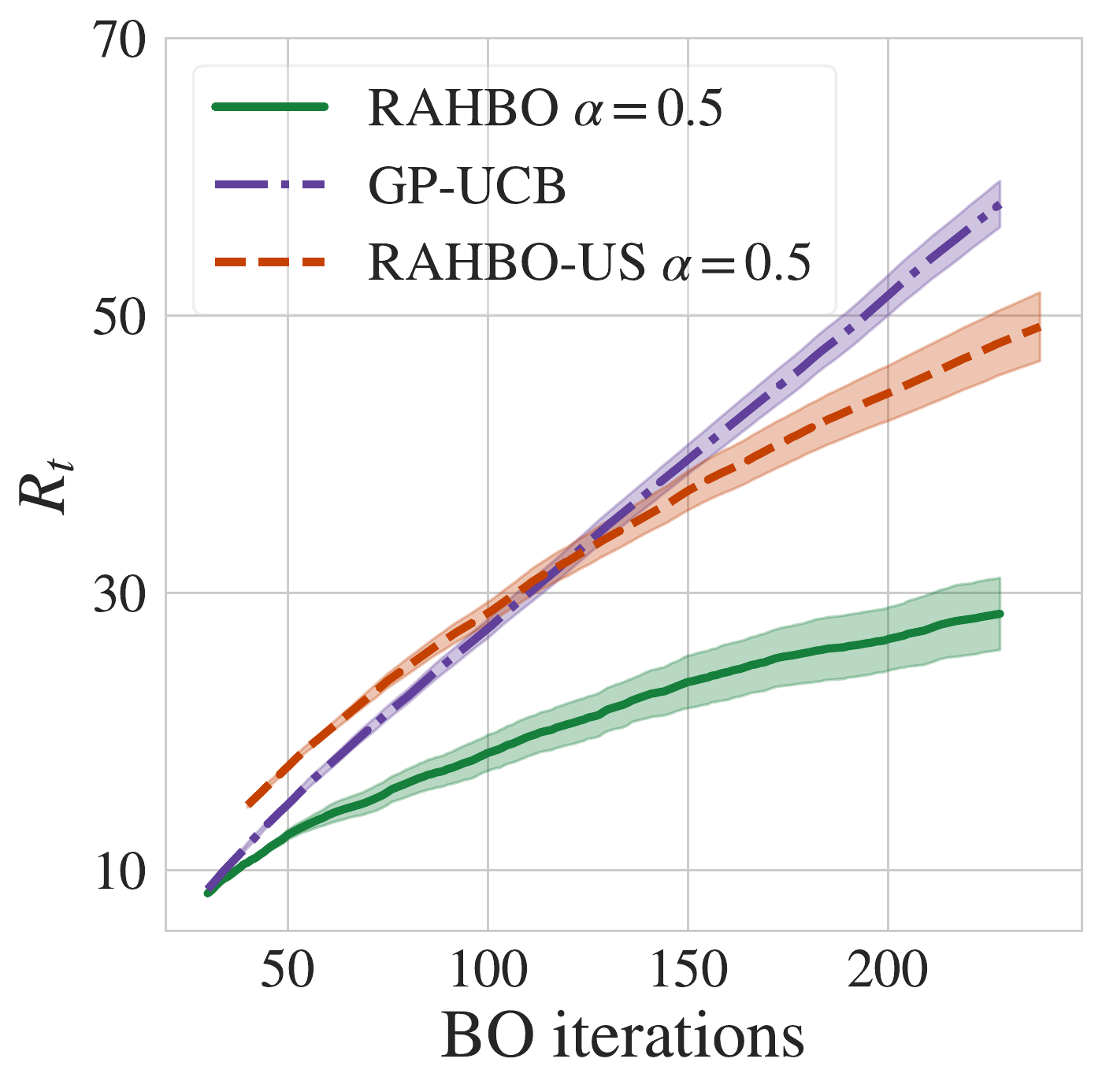}
    \label{fig:swissfel_cumregret_a}
    }
  \subfloat[Cum. regret ($\alpha=1$)]{
    \includegraphics[width=0.26\linewidth]{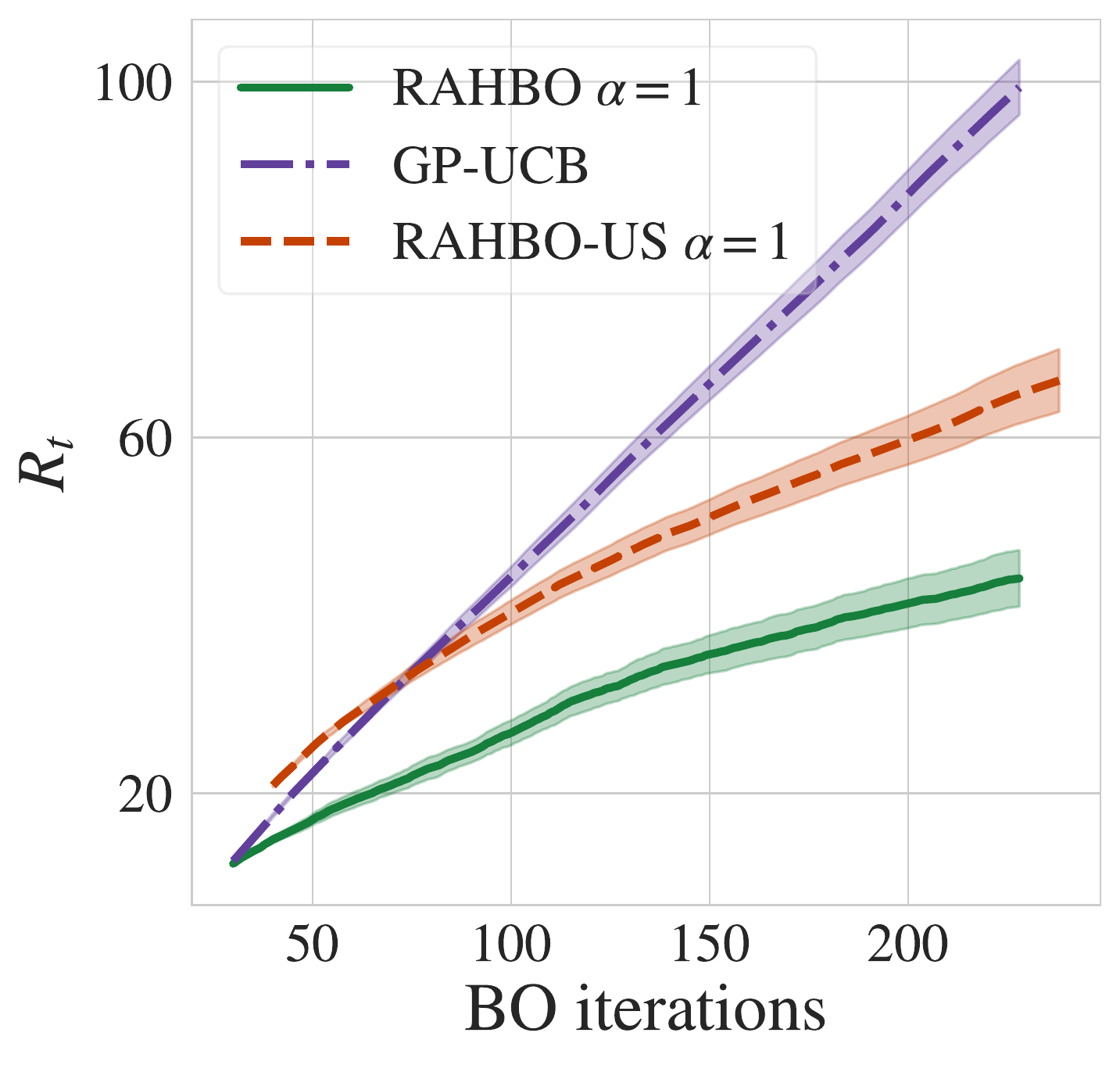}
    \label{fig:swissfel_cumregret_b}
    }
    \caption{Experimental results for \swissfel:  \textbf{(a)} \looseness -1 Distributions of $f(x)$ and $\rho^2(x)$ for \textit{all points} queried during the optimization.  \gpucb queries points with higher noise (but not necessarily high return $f$) in contrast to the risk-averse methods.  \textbf{(b)} Mean $f(\xreported)$ and variance $\rho^2(\xreported)$ at the \textit{reported} $\xreported = \argmax_{x_t}\mathrm{lcb}_T(x_t)$: for each method, we repeat BO experiment 15 times (separate points) and plot corresponding standard deviation error bars. \raucb reports solutions with reasonable mean-variance tradeoff, while \gpucb produces solutions with high mean value but also high noise variance.   \textbf{(c-d)}  Cum. regret for $\alpha  = 0.5$ and $\alpha = 1$ (see more in \cref{app:swissfel}). \looseness-1}
\label{fig:swissfel_res}
\end{figure}

\begin{figure}[H]
        \hspace{-0.8cm}
        \subfloat[Branin benchmark]{
    \includegraphics[width=0.30\linewidth]{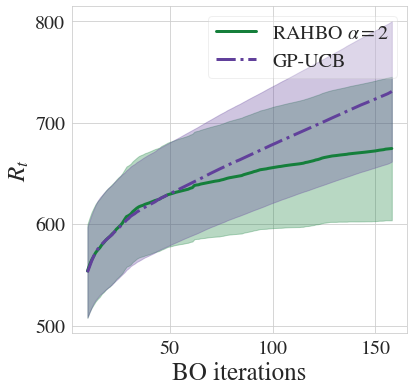}
        \label{fig:branin_cumregret}
        }
    \subfloat[RF Tuning]{
    \includegraphics[width=0.30\linewidth]{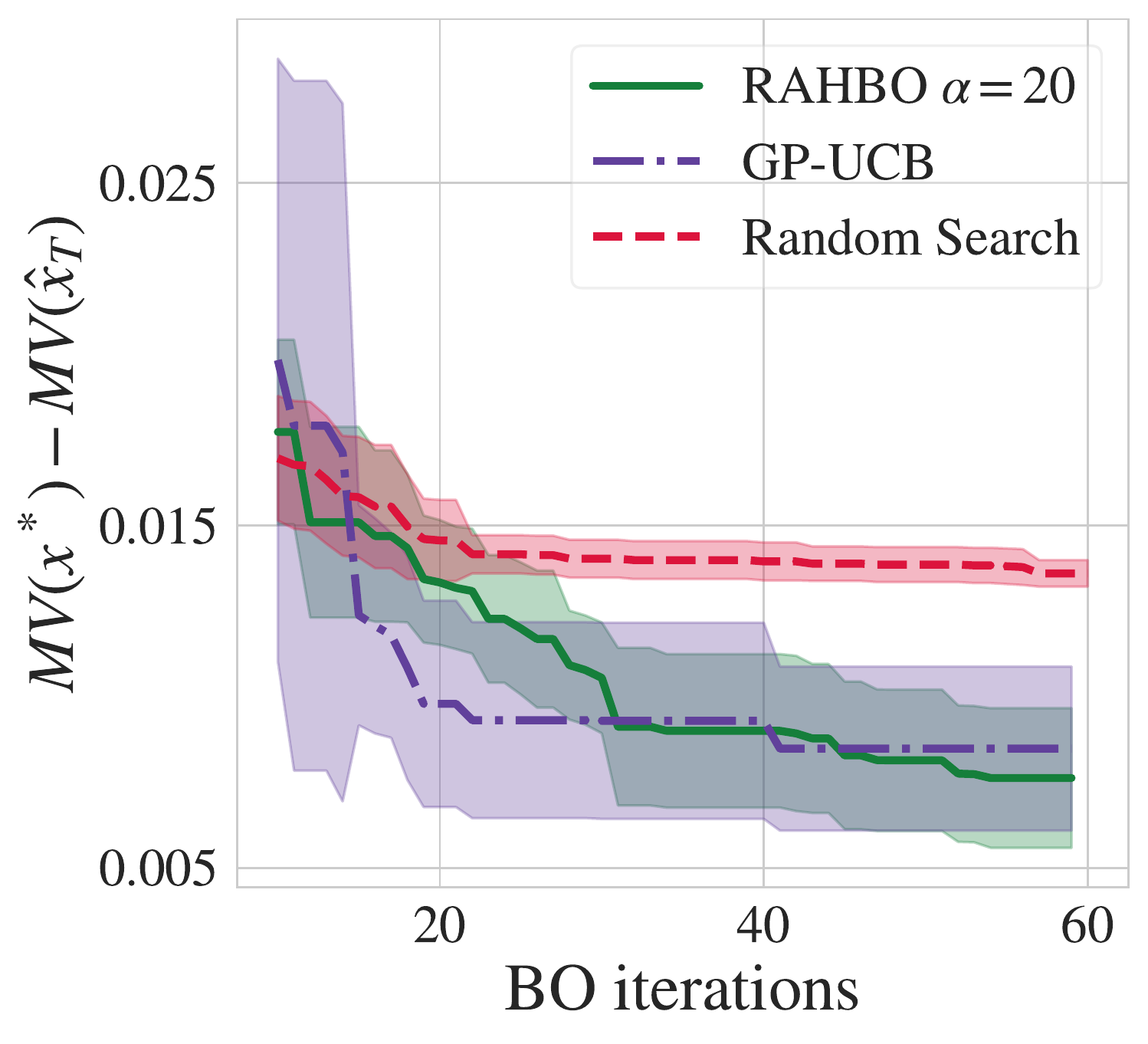}
        \label{fig:rf_mv_10}
        }
    \subfloat[RF Tuning]{
        \includegraphics[width=0.30\linewidth]{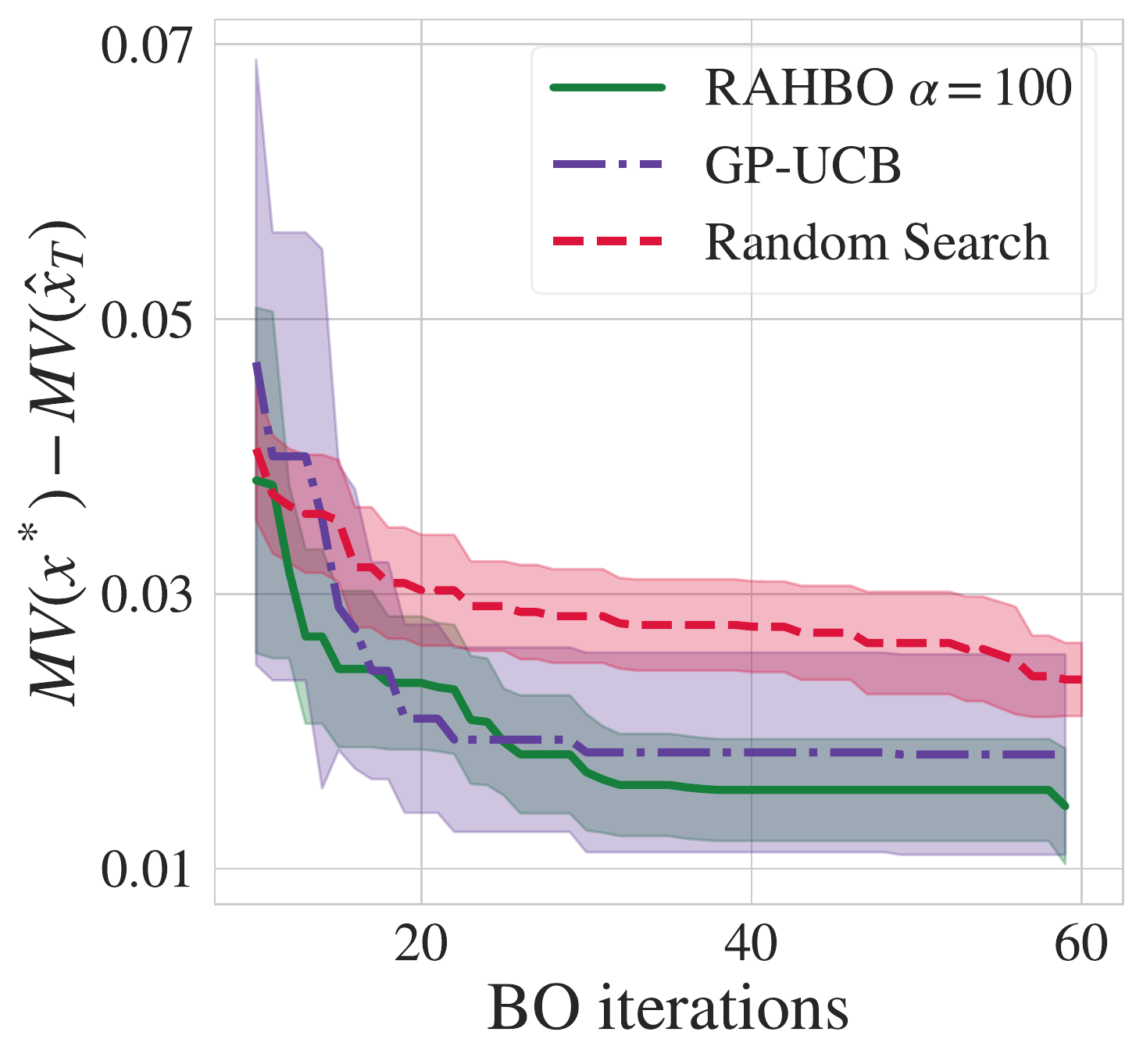} 
        \label{fig:rf_mv_100}
        }
    \caption{{\bf Branin}: (a) Cumulative regret. {\bf Random Forest}: (b-c) Simple regret for the reported $\xreported = \argmax_{x_t}MV(x_t)$ for (b) $\alpha=20$ and (c) $\alpha=100$. While both methods have comparable mean, \raucb has consistently lower variance.} 
    
\end{figure}

We report our comparison in \Cref{fig:swissfel_res} where we also assess the effect of varying the coefficient of absolute risk tolerance $\alpha$. We use 30 points to initialize the baselines and then perform 200 acquisition rounds. We repeat each experiment 15 times. In \Cref{fig:swissfel_hist} we plot the empirical frequency of the true (unknown to the methods) values $f(x_t)$ and $\rho^2(x_t)$ at the inputs $x_t$ acquired by the methods. The empirical frequency for $\rho^2(x)$ illustrates the tendency of risk-neutral \gpucb to query points with higher noise, while risk-averse achieves substantially reduced variance and minimal reduction in mean performance.
Sometimes, risk-neutral \gpucb also fails to succeed in querying points with the highest $f$-value. That tendency results in lower cumulative regret for \raucb in \Cref{fig:swissfel_cumregret_a,fig:swissfel_cumregret_b}. We also compare the performance of the reporting rule from \cref{col_report} in \Cref{fig:swissdel_reported}, where we plot error bars with standard deviation both for $f(\xreported)$ and $\rho^2(\xreported)$ at the reported point $\xreported$. As before, \raucb drastically reduces the variance compared to 
\gpucb, while having only  slightly lower mean performance.
Additional results are presented in \Cref{fig:app_swissfel_res} in Appendix.

\textbf{Random Forest tuning \ \ } \looseness -1 BO is widely used by cloud services for tuning machine learning hyperparameters and the resulting models might be then used in high-stakes applications such as credit scoring or fraud detection. In k-fold cross-validation, the average metric over the validation sets is  optimized -- a canonical example of the \emph{repeated experiment setting} that we consider in the paper. High across-folds variance is a practical problem \cite{makarova2021overfitting} where the mean-variance approach might be beneficial.

In our experiment, we tune hyperparameters of a random forest classifier (RF) on a dataset of fraudulent credit card transactions \cite{bookfrauddetection}.\footnote{\url{https://www.kaggle.com/mlg-ulb/creditcardfraud}}
 ~It consist of ~285k transactions with 29 features (processed due to confidentiality issues) that are distributed over time, and only 0.2\% are fraud examples (see Appendix for more details). The search space for the RF hyperparameters is also provided in the Appendix. We use the balanced accuracy score and 5 validation folds, i.e., $k=5$, and each validation fold is shifted in time with respect to the training data. We seek not only for high performance \emph{on average} but also for low variance across the validation folds that have different time shifts with respect to the training data. 
 
 We initialize the algorithms by selecting 10 hyperparameter settings and keep these points the same for all algorithms. We use Mat\'ern 5/2 kernels with Automatic Relevance Discovery (ARD) and normalize the input features to the unit cube. The number of acquisition rounds in one experiment is 50 and we repeat each experiment 15 times. We demonstrate our results in \Cref{fig:rf_mv_10,fig:rf_mv_100} where we plot mean $\pm$ 2 standard errors. While both \raucb and \gpucb perform comparable in terms of the mean error, its standard deviation for \raucb is smaller.

\section{Conclusion}\label{sec:conclusion}
In this work, we generalize Bayesian optimization to the risk-averse setting and propose \raucb algorithm aiming to find an input with both large expected return and small input-dependent noise variance. Both the mean objective and the variance are assumed to be unknown a~priori and hence are estimated online. \raucb is equipped with theoretical guarantees showing (under reasonable assumptions) sublinear dependence on the number of evaluation rounds $T$ both for cumulative risk-averse regret and $\epsilon$-accurate mean-variance metric. 
The empirical evaluation of the algorithm on synthetic benchmarks and hyperparameter tuning tasks demonstrate promising examples of heteroscedastic use-cases benefiting from \raucb. 
\looseness=-1


\vspace{6pt}
\textbf{Acknowledgements} \\\\
This research has been gratefully supported by NCCR Automation grant 51NF40 180545, by ERC under the European Union’s Horizon grant 815943, SNSF grant 200021\_172781, and ETH Z{ü}rich Postdoctoral Fellowship 19-2 FEL-47. 
The authors thank Sebastian Curi, Mojm\'{i}r Mutn\'{y} and Johannes Kirschner as well as the anonymous reviewers of this paper for their helpful feedback.


\bibliographystyle{abbrv}
\bibliography{paper}

\newpage
\appendix

\section{\Large{Appendix} \\
\textbf{\large{Risk-averse Heteroscedastic Bayesian Optimization \\
\normalsize{\normalfont{(Anastasiia Makarova, Ilnura Usmanova, Ilija Bogunovic, Andreas Krause)}}}}}
\label{sec:appendix}

\subsection{Details on Proposition 1} 
We first provide the proof of Proposition~\ref{proposition_known} for cumulative risk-averse regret \cref{eq:risk_averse_cumulative_regret} with known variance-proxy $\rho^2(\cdot)$ (see \cref{def:subG}) (\cref{sec:proof_known_rho}). We further provide data-independent bounds for $\beta_T$ (\cref{sec:beta_bound}) and maximum information gain $\gamma_T$ (\cref{sec:gamma_bound}) that together conclude the proof for sub-linear on $T$ regret guarantees for most of the popularly used kernels. 

\subsubsection{Proof Proposition 1} \label{sec:proof_known_rho}
\textbf{Proposition 1.}\textit{
   Consider any $f\in \mathcal{H}_{\kappa}$ with $\|f\|_{\kappa} \leq \mathcal{B}_f$ and sampling model from \cref{eq:observational_model} with known variance-proxy $\rho^2(x)$.  Let $\lbrace \beta_t\rbrace_{t=1}^T$ be set as in \cref{lemma:kirschner} with $\lambda=1$. 
Then, with probability at least $1-\delta$, \raucb attains cumulative risk-averse regret $R_T = \mathcal{O}\big(\beta_T \sqrt{T  \gamma_T(\rhomax^2 +1) }\big)$.
}

\begin{proof}
The main steps of the proof are as follows: In \textit{Step 1}, we derive the upper and the lower confidence bounds, $\ucb^{\mv}_t(x_t)$ and $\lcb^{\mv}_t(x_t)$, on $\mv(x_t)$ at iteration $t$. In \textit{Step 2}, we bound the instantaneous risk-averse regret $r(x_t):= \mv(x^*) - \mv(x_t)$. In \textit{Step 3}, we derive mutual information $I(y_{1:T}, f_{1:T})$ in case of the heteroscedastic noise. In \textit{Step 4}, we bound the sum of variances $\sum_{t=1}^T\sigma_{t-1} (x_t)$ via mutual information $I(y_{1:T}, f_{1:T})$. In \textit{Step 5}, we bound the cumulative regret $R_T = \sum_{t=1}^Tr(x_t)$ based on the previous steps.

\textbf{\textit{Step 1}}: 
\textit{On the confidence bounds for $\mv(x)$.}\\
In case of known variance-proxy $\rho^2(x)$, the confidence bounds for $\mv(x)$ at iteration $t$ can be directly obtained based on the posterior $\mu_t(x)$ and $\sigma_t(x)$ for $f(x)$ defined in \cref{eq:gp_mean_hetero,eq:gp_var_hetero}. Particularly, for $\beta_t = \beta_t(\delta)$ defined in \cref{def:beta},
$\Pr \big\{ \mathrm{lcb}_t^{\mv}(x)\leq \mv(x)\leq \mathrm{ucb}_t^{\mv}(x) ~ \forall x \in \mathcal X, \forall t \geq 0 \big\} \geq 1-\delta$  
with the confidence bounds:
\begin{align}
    & \mathrm{lcb}^{\mv}_t(x) := \mu_{t-1}(x) - \beta_t \sigma_{t-1}(x) - \alpha \rho^2(x)  \label{eq:lcb_g},\\ 
    & \mathrm{ucb}^{\mv}_t (x) := \mu_{t-1}(x) + \beta_t \sigma_{t-1}(x) - \alpha \rho^2(x) \label{eq:ucb_g}.
\end{align}

\textbf{\textit{Step 2}}: \textit{On bounding the  instantaneous risk-averse regret $r_t(x)$.} We have
\begin{align*}
    r(x_t) &= {\mv}(x^*) - {\mv}(x_t) 
    \\
    & \leq \mathrm{ucb}^{\mv}_t (x^*) - \mathrm{lcb}^{\mv}_t (x_t) \\ 
    & \leq \mathrm{ucb}^{\mv}_t (x_t) - \mathrm{lcb}^{\mv}_t (x_t)
    = 2\beta_t \sigma_{t-1}(x_{t}),
\end{align*}
where the first inequality is due to the definition of confidence bounds, the second is due to the acquisition strategy $x_{t} \in \argmax_{x\in\X} \ucb^{\mv}_t (x);$ and the equality further expands $\lcb^{\mv}_t(x)$ and $\ucb^{\mv}_t(x).$
Thus, the cumulative regret can be bounded as follows:
\begin{align}R_T = \sum_{t=1}^{T}r(x_t) \leq \sum_{t=1}^{T} 2\beta_t \sigma_{t-1}(x_{t})\leq 2\beta_T\sum_{t=1}^{T}  \sigma_{t-1}(x_{t}),\label{eq:R_known}
\end{align}
where the last inequality holds since $\{\beta_t\}_{t=1}^T$ is a non-decreasing sequence. 

\textbf{\textit{Step 3}}: \textit{ On mutual information $I(y_{1:T}, f_{1:T})$ and maximum information gain $\gamma_T$.}\\
Mutual information $I(y_{1:T}, f_{1:T})$ between the vector of evaluations $y_{1:T} \in \R^T$ at points $A=\{x_t\}_{t=1}^T$ and $f_{1:T}= [f(x_1), \dots, f(x_T)]^\top$ is defined by
   $$I(y_{1:T}, f_{1:T}) = H(y_{1:T}) - H(y_{1:T}|f_{1:T}), $$ where $H(\cdot)$ denotes entropy. 
   Under the modelling assumptions $f_{1:T} \sim \mathcal N(0, \lambda ^{-1}K_T)$ and $\xi_{1:T} \sim \mathcal N(0, \Si_T)$ for the noise $\xi_{1:T} = [\xi(x_1), \dots, \xi(x_T)]^\top$, the measurements are distributed as $y_{1:T}  \sim \mathcal N(0, \lambda ^{-1}K_T +  \Si_T)$ and $y_t|y_{1:t-1} \sim \mathcal{N} (\mu_{t-1}(x_t), \rho^2(x_t) + \sigma^2_{t-1}(x_t))$, where $\sigma^2_{t-1}(\cdot)$ is defined in \cref{eq:gp_var_hetero}. Hence, the entropy of each new measurement $y_t$ conditioned on the previous history $y_{1:t-1}$ is: \looseness-1
\begin{align*}
    H(y_t|y_{1:t-1})& = \frac{1}{2} \ln \left(2\pi e \left(\rho^2(x_t) + \sigma^2_{t-1}(x_t)\right)\right) \\
    &
     = \frac{1}{2} \ln \bigg( 2\pi e \rho^2(x_t) \left( 1+ \frac{\sigma^2_{t-1}(x_t)}{  \rho^2(x_t)}\right) \bigg) \\
    & = \frac{1}{2} \ln \bigg(2\pi e   \rho^2(x_t) \bigg)
    + \frac{1}{2} \ln \bigg( 1+ \frac{\sigma^2_{t-1}(x_t)}{   \rho^2(x_t)}\bigg), \\
     H(y_{1:T}) & = \sum_{t=1}^T H(y_t|y_{1:t-1}) = \frac{1}{2}  \sum_{t=1}^T\ln \bigg(2\pi e   \rho^2(x_t) \bigg) 
                                                + \frac{1}{2}\sum_{t=1}^T \ln \bigg( 1+ \frac{\sigma^2_{t-1}(x_t)}{   \rho^2(x_t)} \bigg),\\
     H(y_{1:T}|f_{1:T}) &= \sum_{t=1}^T H(y_t|f_t) = \frac{1}{2} \sum_{t=1}^T \ln (2\pi e   \rho^2(x_t)).
    \end{align*}
    Therefore, the information gain for $y_{1:T}$ is:
    \begin{align}
    &I(y_{1:T}, f_{1:T}) = H(y_{1:T}) - H(y_{1:T}|f_{1:T}) = \frac{1}{2}\sum_{t=1}^T \ln \bigg( 1+ \frac{\sigma^2_{t-1}(x_t)}{   \rho^2(x_t)} \bigg). \label{app:eq_information_gain_known_rho}
\end{align}
Then, by definition of maximum information gain:
\begin{align}
\gamma_T := \underset{A\subset\mathcal{X},\; |A| = T}{\max} \; I(y_{1:T}, f_{1:T}) \geq \frac{1}{2}\sum_{t=1}^T \ln \bigg( 1+ \frac{\sigma^2_{t-1}(x_t)}{   \rho^2(x_t)} \bigg).  \label{eq:info_gain_app}
\end{align}

\textbf{\textit{Step 4}}: \textit{On bounding $\sum_{t=1}^T\sigma_{t-1} (x_t)$.} \\

\begin{align}
    \sum_{t=1}^T\sigma_{t-1}(x_t) 
    & =   \sum_{t=1}^T \frac{\rho(x_t)}{\rho(x_t)}\sigma_{t-1}(x_t) \leq 
    \sqrt{T \sum_{t=1}^T  \rho^2(x_t) \frac{\sigma^2_{t-1}(x_t)}{ \rho^2(x_t)}}\nonumber \\
    & \leq 
     \sqrt{ T   \sum_{t=1}^T \frac{1}{\ln(1+\rho^{-2}(x_t))} \ln \bigg( 1+ \frac{\sigma_{t-1}^2(x_t) }{ \rho^2(x_t)} \bigg)}\nonumber \\
      & \leq 
     \sqrt{ \frac{ 2T }{\ln(1+\rhomax^{-2})} \underbrace{\frac{1}{2}\sum_{t=1}^T \ln \bigg( 1+ \frac{\sigma_{t-1}^2(x_t) }{ \rho^2(x_t)} \bigg)}_{\mathrm{mutual~ information~ \cref{app:eq_information_gain_known_rho}}}},\label{eq:bound_variances_known_rho}
\end{align}
where the first inequality follows from the Cauchy-Schwarz inequality. The second one is due to the fact that
for any $s^2\in[0,\rho^{-2}(x_t)]$ we can bound $s^2 \leq \frac{\rho^{-2}(x_t)}{\ln(1+\rho^{-2}(x_t))} \ln(1+s^2)$, that also holds for $s^2:=\rho^{-2}(x_t)\sigma^2_{t-1}(x_t) $ since $\rho^{-2}(x_t)\sigma^2_{t-1}(x_t) \leq \rho^{-2}(x_t)\lambda^{-1}\kappa(x_t,x_t) \leq \rho^{-2}(x_t)$ for $\lambda \geq 1$. 
The third inequality is due to $\rho(x) \in [\rhomin, \rhomax]$.

\textbf{\textit{Step 5}}: \textit{Bounding risk-averse cumulative regret  $R_T=\sum_{t=1}^{T}r(x_t)$.}\\
Combining the previous three steps together: \cref{eq:R_known}, \cref{eq:info_gain_app}, and \cref{eq:bound_variances_known_rho} we finally obtain:
\begin{align*}
    & R_T 
    \leq \sum_{t=1}^{T} 2\beta_t \sigma_{t-1}(x_{t}) \leq 2\beta_T \sum_{t=1}^{T}  \sigma_{t-1}(x_{t}) 
    \leq 2\beta_T \sqrt{\frac{2T }{\ln(1+\rhomax^{-2})}\gamma_T}
\end{align*}
    Also, note that for any $\alpha \geq 0$ the bound  $\ln(1+\alpha) \geq \frac{\alpha}{1+\alpha}$ holds, thus $\frac{1}{\ln(1+\rhomax^{-2})} \leq \frac{1+\rhomax^{-2}}{\rhomax^{-2}} = \rhomax^{2} + 1.$ Therefore, the cumulative regret can be also bounded as
    $R_T = O(\beta_T\sqrt{T\gamma_T(\rhomax^2 + 1)}).$
\end{proof}

\subsubsection{Bounds for $\beta_T$}  \label{sec:beta_bound}
We provide the bounds for the data-dependent $\beta_T$ that appear in the regret bound (see \cref{def:beta}). Following our modelling assumptions $f_{1:T} \sim \mathcal N (0, \lambda^{-1}K_T)$ and $\xi_{1:T} \sim \mathcal N (0, \Sigma_T)$, the information gain $I(y_{1:T}, f_{1:T}) = H(y_{1:T}) - H(y_{1:T}|f_{1:T}) $ is given as follows:

\begin{align}
    I(y_{1:T}, f_{1:T})  = \underbrace{\frac{1}{2} \ln \big(\det (2\pi e (\lambda^{-1}K_T + \Sigma_T) ) \big)}_{H(y_{1:T})} - \underbrace{\frac{1}{2} \ln \big(\det (2\pi e \Sigma_T) \big)}_{H(y_{1:T}|f_{1:T})} = \frac{1}{2} \ln\left( \frac{\det(K_T +  \lambda\Sigma_T)}{\det (\lambda\Sigma_T)}\right).
\end{align}

By definition then $\gamma_T = \underset{A\subset\mathcal{X},\; |A| = T}{\max} \; I(y_{1:T}, f_{1:T}) \geq \frac{1}{2} \ln\left( \frac{\det(K_T +  \lambda\Sigma_T)}{\det (\lambda\Sigma_T)}\right)$.
On the other hand, $\beta_T$ defined in Lemma 1 can be expanded in a data-independent manner as follows: 

\begin{align}
\beta_T &:= \sqrt{2\ln \bigg( \frac{\det(\lambda \Sigma_T +K_T)^{1/2}}{\delta\det(\lambda \Sigma_T)^{1/2}}}
        \bigg) + \sqrt{\lambda}\|f\|_{\kappa}   \nonumber \\ &=  \sqrt{2\ln \frac{1}{\delta} + \ln \left( \frac{\det(\lambda \Sigma_T +K_T)}{\det(\lambda \Sigma_T)}\right)} +  \sqrt{\lambda} \|f\|_{\kappa}  
        \leq \sqrt{2\ln \frac{1}{\delta} + \gamma_T} + \sqrt{\lambda}B_f.
\end{align}

\subsubsection{Bounds for $\gamma_T$} \label{sec:gamma_bound}
Here, we show the relation between the information gains under heteroscedastic and homoscedastic noise. Note that for the latter the upper bounds are widely known, e.g., \cite{srinivas10}. To distinguish between the maximum information gain for heteroscedastic noise with variance-proxy $\rho^2(x)$ and the maximum information gain for homoscedastic noise with fixed variance-proxy $\sigma^2$, we denote them as  $\gamma_T^{\rho_x}$ and  $\gamma_T^{\sigma}$ respectively. Recall that $\varrho^2(\cdot) \in [\smash{\underline{\varrho}^2}, \bar{\varrho}^2]$ for some constant values $\bar{\varrho}^2 \geq \smash{\underline{\varrho}^2} > 0$. 

Below, we show that $\gamma_T^{\rho_x} \leq  \gamma_T^{\sigma}  \frac{\bar{\varrho}^2}{\smash{\underline{\varrho}^2}} $ with $\sigma^2$ set to  $\bar{\varrho}^2 $,  that only affects the constants but not the main scaling (in terms of $T$) of the known bound for the homoscedastic maximum information gain.

\begin{align}
    & \gamma_T^{\rho_x}  
        & \overset{\tiny \circled{1}}{=} 
            \underset{A\subset\mathcal{X},|A| = T}{\max} \frac{1}{2}\sum_{t=1}^T \ln \bigg( 1+ \frac{\sigma^2_{t-1}(x_t | \rho^2(x_t))}{ \rho^2(x_t)} \bigg) 
        &\overset{\tiny \circled{2}}{\leq}
            \underset{A\subset\mathcal{X},|A| = T}{\max} \frac{1}{2}\sum_{t=1}^T \ln \bigg( 1+ \frac{\sigma^2_{t-1}(x_t | \rhomax^2)}{ \rhomin^2} \bigg)  \\
    &&\overset{\tiny \circled{3}}{=} 
        \underset{A\subset\mathcal{X},|A| = T}{\max} \frac{1}{2}\sum_{t=1}^T \ln \bigg( 1+ \frac{\rhomax^2}{\rhomin^2} \frac{\sigma^2_{t-1}(x_t | \rhomax^2)}{\rhomax^2} \bigg) 
        &\overset{\tiny \circled{4}}{\leq}
            \underset{A\subset\mathcal{X},|A| = T}{\max} \frac{1}{2}\sum_{t=1}^T \frac{\rhomax^2}{\rhomin^2} \ln \bigg( 1+ \frac{\sigma^2_{t-1}(x_t | \rhomax^2)}{\rhomax^2} \bigg)  \\
    &&\overset{\tiny \circled{5}}{=}
        \underset{A\subset\mathcal{X}, |A| = T}{\max} \frac{\rhomax^2}{\rhomin^2} \frac{1}{2}\sum_{t=1}^T \ln \bigg( 1+ \frac{\sigma^2_{t-1}(x_t | \sigma^2)}{\sigma^2} \bigg)  
        & = \frac{\rhomax^2}{\rhomin^2} \gamma_T^{\sigma}, \label{app:eq_info_gain_bound}
\end{align}

where  $\tiny\circled{1}$ follows from \cref{app:eq_information_gain_known_rho}. In $\tiny\circled{2}$, we lower bound the denominator $\rho^2(x_t)$ and upper bound the numerator $\sigma^2_{t-1}(x_t)$ (due to monotonicity w.r.t.~noise variance, i.e., $\sigma^2_{t-1}(x_t| \Sigma_t) \leq \sigma^2_{t-1}(x_t| \bar{\varrho}^2 \mathbf I_t)$). In $\tiny\circled{3}$,  we multiply by $1=\bar{\varrho}^2/\bar{\varrho}^2$. In $\tiny\circled{4}$  we use Bernoulli inequality since $\bar{\varrho}^2/\smash{\underline{\varrho}^2} \geq 1$. The obtained expression can be interpreted as a standard information gain for homoscedastic noise and, particularly, with the variance-proxy $\sigma^2$ set to $ \bar{\varrho}^2 $ due to $\tiny\circled{5}$. Finally, the upper bounds on $\gamma_T^{\sigma}$ typically scale sublinearly in $T$ for most of the popularly used kernels \cite{srinivas10}, e.g, for linear kernel $\gamma_T = \mathcal O(d\log T)$, and for squared exponential kernel $\gamma_T = \mathcal O(d(\log T)^{d+1})$.


\subsection{Tighter bounds for the variance-proxy $\rho^2_{\eta}(x)$.}\label{sec:varproxy_for_var}
Assumption \ref{asm:subG_known} states that noise $\eta(x)$ from \cref{eq:sample_variance_eval} is $\rho^2_{\eta}(x)$-sub-Gaussian with variance-proxy $\rho^2_{\eta}(x)$ being known. In practice, $\rho^2_{\eta}(x)$ might be unknown. 
Here, we describe a way to estimate $\rho_{\eta}^2(x)$ under the following two assumptions: the evaluation noise $\xi(x)$ is strictly sub-Gaussian (that is already reflected in the Assumption \ref{asm:ssubG})  and the noise $\eta(x)$ of variance evaluation is also strictly sub-Gaussian, that is, $\mathbb Var [\eta(x)] = \rho^2_{\eta}(x)$ and $\mathbb Var [\xi(x)] = \rho^2(x)$. 

\textit{(i) Reformulation of the sample variance.} We first rewrite the sample variance defined in \cref{eq:emp_mean_and_var} as the average over squared differences over all pairs $\{y_1(x), \dots, y_k(x)\}$:

\begin{align}
    \samplevar &
    \overset{\tiny \circled{1}}{=} 
         \frac{1}{2k(k-1)} \sum_{i=1}^k \sum_{j=1}^k (y_i(x) - y_j(x) )^2 
    \overset{\tiny \circled{2}}{=} 
        \frac{1}{2k(k-1)} \sum_{i=1}^k \sum_{j=1}^k (\xi_i(x) - \xi_j(x))^2, \label{eq:sample_var_reformulated}
\end{align}
where $\tiny\circled{2}$ is due to $y_i(x) = f(x) + \xi_i(x)$, and $\tiny\circled{1}$ is equivalent to the \cref{eq:emp_mean_and_var} due to the following:
\begin{align}
    &\frac{1}{2k(k-1)} \sum_{i=1}^k \sum_{j=1}^k (y_i - y_j)^2 = \frac{1}{2k(k-1)} \sum_{i=1}^k \sum_{j=1}^k (y_i - \hat{m}_{\krepeat} + \hat{m}_{\krepeat} - y_j)^2  \\
    &=
        \frac{1}{2\krepeat(\krepeat - 1)} \sum_{i=1}^{\krepeat}\sum_{j=1}^{\krepeat}\big[\big(y_i  - \hat{m}_{\krepeat}\big)^2 + \big(y_j - \hat{m}_{\krepeat}\big)^2 - 2\big(y_i  - \hat{m}_{\krepeat}\big)\big(y_j - \hat{m}_{\krepeat}\big)\big]  \nonumber \\
    &=\frac{1}{2\krepeat(\krepeat - 1)} \sum_{i=1}^{\krepeat}\sum_{j=1}^{\krepeat}\big[\big(y_i - \hat{m}_{\krepeat}\big)^2 + \big(y_j- \hat{m}_{\krepeat}\big)^2 \big]  \nonumber \\
    &= \frac{1}{\krepeat - 1} \sum_{i=1}^{\krepeat}(y_i  - \hat{m}_{\krepeat})^2.  
\end{align}

\textit{(ii) Variance of the sample variance $\mathbb Var[\samplevar]$.} In \cref{eq:sample_var_reformulated}, we show that sample variance can be written in terms of the noise $\xi(x)$. In \cite{benhamou2018properties} (see Eq. (37)), it is shown that for i.i.d observations $\{\xi_1(x), \dots, \xi_k(x)\}$, sampled from a distribution with the 2nd and 4th central moments $\mathbb Var[\xi(x)]$ and $ \mu_4(x) = \E [\xi^4(x)]$, respectively, the variance of the sample variance can be computed as follows: \looseness-1
\begin{align*}
     \mathbb Var[\samplevar] = \E [\big(\samplevar\big)^2] - \E[\samplevar]^2 = \frac{\mu_4(x)}{k} - \frac{(k-3) \mathbb Var^2[\xi(x)]}{k(k-1)}.
\end{align*}
Since $\xi(x)$ is strictly $\rho(x)$--sub-Gaussian, the latter can be further adapted as 
\begin{align*}
     \mathbb Var[\samplevar] = \frac{\mu_4(x)}{k} - \frac{(k-3)\rho^4(x)}{k(k-1)}.
\end{align*}


\textit{(iii)} Due to $\eta(x)$ being  strictly sub-Gaussian, i.e., $\rho^2_{\eta}(x) =  \mathbb Var[\eta(x)] =\mathbb Var[\samplevar]$, the derivation above also holds for the variance-proxy $\rho^2_{\eta}(x)$:

$$
\rho^2_{\eta}(x) = \frac{\mu_4(x)}{k} - \frac{(k-3)\rho^4(x)}{k(k-1)}.
$$

\textit{(iv) Bound 4th moment $\mu_4(x)$.}  The 4th moment $\mu_4(x)$ can expressed in terms of the distribution kurtosis that is bounded under our assumptions. 
Particularly, \emph{kurtosis} $\mathrm{Kurt}[\xi]:=  \frac{\E[(\xi - \E[\xi])^4]}{\mathbb Var^2(\xi)}
$ is measure that identifies the tails behaviour of the distribution of $\xi$; $\mathrm{Kurt}(\xi) = 3$ for normallly distribute $\xi$ and $\mathrm{Kurt}(\xi) \leq 3$  for strictly sub-Gaussian random variable $\xi$ (see \citep{Arbel2019OnSS}). This implies 
$$\mu_4 (x) = \mathrm{Kurt}\big(\xi(x)\big)  \rho^4(x) \leq 3\rho^4(x).$$


\textit{(v) Bound variance-proxy.} There
\begin{align*}
    \rho^2_{\eta}(x)  
    \leq \frac{3(k-1)\rho^4(x) - (k-3)\rho^4(x)}{k(k-1)}
    = \frac{3 k-3- k+3}{k(k-1)}\rho^4(x)
    =\frac{2\rho^4(x)}{k-1}. 
\end{align*}
In case of the known bound $\rhomax^2 \geq \rho^2(x)$, we bound the unknown $\rho^2_{\eta}(x)$ as follows:
\begin{align*}
    \rho^2_{\eta}(x) 
    &\leq 
    \frac{2\rhomax^4}{k-1}. 
\end{align*}
\subsection{Method details: GP-estimator of variance-proxy $\rho^2$ }\label{app:method}
According to the Assumption \ref{asm:subG_known}, variance-proxy $\rho^2 \in \mathcal{H}_{\kappa^{var}}$ is smooth, and $\eta(x) = \hat{s}_{\krepeat}^2(x) - \rho^2(x)$ is $\rho_{\eta}(x)$-sub-Gaussian with known variance-proxy $\rho^2_{\eta}(x)$. In this case, confidence bounds for $\rho^2(x)$ follow the ones derived in  \cref{lemma:kirschner} with $\beta^{var}_t$ based on $\Sigma_t^{var}$. Particularly, we collect noise variance evaluations $\{x_t, \hat{s}_{\krepeat}(x_t)\}_{t=0}^T.$
Then the estimates for $\mu^{\varproxy}_T(x)$ and $\sigma^{\varproxy}_T(x)$ for $\rho^2$ follow the corresponding estimates for $f(x)$. Particularly, 


\begin{align}
    \mu^{\varproxy}_t(x) &= \kappa^{\varproxy}_t(x)^T(K^{\varproxy}_t + \lambda \Si^{\varproxy}_t)^{-1} \hat{s}_{1:t} \label{eq:gp_mean_hetero_1},\\
     \sigma_t^{\varproxy}(x)^2 &= \frac{1}{\lambda} (\kappa^{\varproxy}(x,x) - \kappa^{\varproxy}_t(x)^\top(K^{\varproxy}_t + \lambda \Si^{\varproxy}_t)^{-1}\kappa^{\varproxy}_t(x)), \label{eq:gp_var_hetero_1}
     \end{align} 

where $\Si^{\varproxy}_t = \text{diag}[\rho_{\eta}^2(x_1), \dots, \rho_{\eta}^2(x_t)]$, $ \kappa^{\varproxy}_t(x) =  [\kappa^{\varproxy}(x_1, x), \dots, \kappa^{\varproxy}(x_t, x) ]^T$ and $(K^{\varproxy}_t)_{i,j} = \kappa^{\varproxy}(x_i, x_j)$. The confidence bounds are then:
\begin{align*}
    \mathrm{ucb}_{t}^{\varproxy}(x) &= \mu^{\varproxy}_{t-1}(x) + \beta_t^{\varproxy} \sigma^{\varproxy}_{t-1}(x)
    \\
    \mathrm{lcb}_t^{\varproxy}(x) &= \mu^{\varproxy}_{t-1}(x) - \beta_t^{\varproxy} \sigma^{\varproxy}_{t-1}(x)
,
\end{align*}
with $\lbrace \beta^{var}_t\rbrace_{t=1}^T$ set according to \cref{lemma:kirschner}.

\subsection{Proof of Theorem~\ref{theorem}} \label{app:proof_unknown_rho}

\begin{customthm}{1}\textit{
Consider any $f\in \mathcal{H}_{\kappa}$ with $\|f\|_{\kappa} \leq \mathcal{B}_f$ and sampling model in \cref{eq:observational_model} with unknown variance-proxy $\rho^2(x)$ that satisfies Assumptions~\ref{asm:ssubG} and \ref{asm:subG_known}. Let $\{x_t\}_{t=1}^T$ denote the set of actions chosen by \raucb (\cref{alg:risk-averse-bo}) over T rounds.
 Set  $\lbrace \beta_t\rbrace_{t=1}^T$ and $\lbrace \beta^{var}_t\rbrace_{t=1}^T$ according to \cref{lemma:kirschner} with $\lambda=1$, $\mathcal{R}^2 = \max_{x\in \mathcal X} \rho^2_{\eta}(x_t)$ and $\rho(\cdot) \in [\rhomin, \rhomax]$.  Then, the risk-averse cumulative regret $R_{T}$ of \raucb is bounded as follows}: 
\begin{align}
& \mathrm{Pr} 
\bigg\{
R_{T}  \leq 
 \beta_T k\sqrt{\frac{2T\hat\gamma_T}{ \ln (1 + k/\rhomax^{2})}} + \alpha \beta_T^{var} k\sqrt{\frac{2T\Gamma_T }{\ln (1 + \mathcal{R}^{-2})}} 
,\quad \forall T \geq 1 \bigg\} \geq 1 - \delta.
\end{align}
\end{customthm}

\textit{Proof.} The main steps of our proof are as follows:  In \textit{Step 1}, we derive the upper and the lower confidence bounds, $\ucb^{\mv}_t(x_t)$ and $\lcb^{\mv}_t(x_t)$, on $\mv(x_t)$ at iteration $t$. In \textit{Step 2}, we bound the instantaneous risk-averse regret $r(x_t):= \mv(x^*) - \mv(x_t)$. In \textit{Step 3}, we derive mutual information both for function and variance-proxy evaluations. In \textit{Step 4}, we bound the sum of variances via mutual information. In \textit{Step 5}, we bound the cumulative regret $R_T = \sum_{t=1}^Tr(x_t)$ based on the previous steps.

\textit{\textbf{Step 1:} On confidence bounds for  $\mv(x)$.}

\textit{(i) On confidence bounds for $\rho^2(x)$.} 
According to  \cref{eq:gp_var_hetero_1}, with probability $1-\delta$ the following confidence bounds hold with $\lbrace \beta^{var}_t\rbrace_{t=1}^T$ set according to \cref{lemma:kirschner}:
\begin{align*} 
    \mathrm{ucb}_t^{\varproxy}(x) &= \mu^{\varproxy}_{t-1}(x) + \beta_t^{\varproxy} \sigma^{\varproxy}_{t-1}(x),
    \\
    \mathrm{lcb}_t^{\varproxy}(x) &= \mu^{\varproxy}_{t-1}(x) - \beta_t^{\varproxy} \sigma^{\varproxy}_{t-1}(x).
\end{align*}
\textit{(ii) On confidence bounds for $f(x)$.}  Here we adapt confidence bounds introduced in \cref{eq:lcb_g}-(\ref{eq:ucb_g})
since \cref{eq:gp_mean_hetero} relies on the unknown variance-proxy  $\rho^2(x)$ incorporated into $\Si_T$. Conditioning on the event that $\rho^2(x)$ is upper bounded by $\mathrm{ucb}_t^{\varproxy}(x) \geq \rho(x)^2$ defined in (i), the confidence bounds for $f$ with probability $1-\delta$ are:
 \begin{align}
    & \mathrm{ucb}^f_t(x) = \mu_{t-1}(x|\hat \Sigma_{t-1}) + \beta_t \sigma_{t-1}(x|\hat \Sigma_{t-1}) \label{eq:ucb_f_unknown_rho}, \\
    & \mathrm{lcb}^f_t(x) = \mu_{t-1}(x|\hat \Sigma_{t-1}) - \beta_t \sigma_{t-1}(x|\hat \Sigma_{t-1}),~  \forall x, t \label{eq:lcb_f_unknown_rho}.
 \end{align}

\textit{(iii) On confidence bounds for $\mv(x)$.} Finally, combining (i) and (ii) and using the union bound, with probability $1-2\delta$, we get $\mathrm{lcb}^{\mv}_t(x) \leq\mv(x)\leq \mathrm{ucb}^{\mv}_t(x)$ with
  \begin{align}
    & \mathrm{ucb}^{\mv}_t(x) = \mathrm{ucb}^f_t(x) - \alpha \mathrm{lcb}^{\varproxy}_t(x), \label{eq:ucb_g_unknown_rho}\\
    & \mathrm{lcb}^{\mv}_t(x) = \mathrm{lcb}^f_t(x) - \alpha \mathrm{ucb}^{\varproxy}_t(x) \label{eq:lcb_g_unknown_rho},~ \forall x, t.
 \end{align} 


\textbf{\textit{Step 2:}} \textit{On bounding the instantaneous regret.}\\
First, we bound instantaneous regret of a single measurement at point $x_t$, but with unknown variance-proxy $\rho^2(x)$ as follows:
\begin{align}
    r_t 
    := \mv(x^*) - \mv(x_t) 
    & 
    \leq
    \mathrm{ucb}^{\mv}_t(x^*) - \mathrm{lcb}^{\mv}_t(x_t) 
    \nonumber \\
    &
    \leq \mathrm{ucb}^{\mv}_t(x_t) - \mathrm{lcb}^{\mv}_t(x_t) 
    \nonumber 
    \\
    & = \mathrm{ucb}^f_t(x_t) - \mathrm{lcb}^f_t(x_t) + \alpha (\mathrm{ucb}^{\varproxy}_t(x_t) - \mathrm{lcb}^{\varproxy}_t(x_t)) \nonumber
    \\
    & = 2 \beta_t \sigma_{t-1}(x_t|\hat \Sigma_{t-1}) + 2 \alpha \beta_t^{\varproxy} \sigma_{t-1}^{\varproxy}(x_t)  \label{eq:regret_bound_ucb_unknown_rho}.
\end{align} 
The second inequality is due to the acquisition algorithm. The last equality is due to the fact that $\ucb^f_t(x) - \lcb^f_t(x) = 2\beta_t\sigma_{t-1}(x_t)$ by definition, as well as $\ucb^{var}_t(x) - \lcb^{var}_t(x) = 2\beta^{var}_t\sigma^{var}_{t-1}(x_t).$

Note that at each iteration $t$ we take $k$ measurements, hence the total number of measurements is $Tk$.  Thus, we can bound the cumulative regret by 
\begin{align} 
R_{T}& 
= \sum_{t=1}^{T} k r(x_t)  
    \leq
        k\sum_{t=1}^{T} 2 \beta_t \sigma_{t-1}(x_t|\hat \Sigma_{t-1}) 
        + k\sum_{t=1}^{T}2 \alpha \beta_t^{\varproxy} \sigma_{t-1}^{\varproxy}(x_t) \nonumber
    \\
    &\leq 2k\beta_T \sum_{t=1}^{T}  \sigma_{t-1}(x_t|\hat \Sigma_{t-1}) +  2k\alpha\beta_T^{var}\sum_{t=1}^{T} \sigma_{t-1}^{\varproxy}(x_t).
\end{align}

\textit{\textbf{Step 3}: On bounding maximum information gain.} \\
We follow the notion of information gain $I(\hat m_{1:T}, f_{1:T})$ computed assuming that $\hat m_{1:T} = [\hat m_k(x_1), \ldots,\hat m_k(x_T)]^T$ with $
    \hat{m}_{\krepeat}(x_t)= \frac{1}{\krepeat} \sum_{i=1}^{\krepeat} y_i(x_t) $ (\cref{eq:emp_mean_and_var}). Under the modelling assumptions $f_{1:T} \sim \mathcal N(0, \lambda ^{-1}K_T)$, and $\hat{m}_{1:T} \sim \mathcal N(f_{1:T}, \text{diag}(\rhomax^2/k))$ 
 with variance-proxy $\rhomax^2/k$, the information gain is:
\begin{align}
    &I(\hat m_{1:T}, f_{1:T}) := \sum_{t=1}^T \frac{1}{2}\ln \bigg ( 1+  \frac{\sigma_{t-1}^2(x_t|\text{diag}(\rhomax^2/k))}{ \rhomax^2/k} \bigg). \label{eq:mutual_information_mean}
\end{align}
We define the corresponding maximum information gain $\hat\gamma_T = \max_{A\subset \X, |A|=T}  I(\hat m_{1:T}, f_{1:T})$ 
\begin{align}\hat\gamma_T := 
\max_{A\subset \X, |A|=T}\sum_{t=1}^T \frac{1}{2}\ln \bigg ( 1+  \frac{\sigma_{t-1}^2(x_t|\text{diag}(\rhomax^2/k))}{ \rhomax^2/k} \bigg).\label{def:hat_gamma}
\end{align}

Analogously, for $\rho(x)$ with the posterior $\mathcal{N}(\mu_t^{\varproxy}(x), (\sigma_t^{\varproxy}(x))^2),$
the information gain is defined as:
\begin{align}
    &I(\hat{s}^2_{1:T}, \rho^2_{1:T}) := \frac{1}{2}\sum_{t=1}^T \ln\left( 1+ \frac{ (\sigma_{t-1}^{\varproxy})^2(x)}{ \rho_{\eta}^2(x_t)}\right).  \label{eq:mutual_information_variance}
\end{align}
Then, the corresponding maximum information gain $\Gamma_T $ is as follows:
\begin{align}
\Gamma_T := \max_{A\subset \X,|A| = T}  I(\hat{s}^2_{1:T}, \rho^2_{1:T}) = \max_{A\subset \X,|A| = T}  \frac{1}{2}\sum_{t=1}^T \ln\left( 1+ \frac{ (\sigma_{t-1}^{\varproxy})^2(x)}{ \rho_{\eta}^2(x_t)}\right),\label{def:Gamma}
\end{align}
where $A$ is again a set of size $T$ with points $\{x_1,\ldots,x_T\}$ .

\textbf{\textit{Step 4}}: \textit{On bounding
$\sum_{t=1}^T\sigma_{t-1}(x_t|\hat \Sigma_{t-1}) $ and $\sum_{t=1}^{T} \sigma_{t-1}^{\varproxy}(x_t)$} \\
We repeat the corresponding derivation for known $\rho^2(x)$, recalling that $\rho^2(x) \leq \rhomax^2, \forall x\in\X$:
\begin{align}
     \sum_{t=1}^T\sigma_{t-1}(x_t|\hat \Sigma_{t-1}) & =   \sum_{t=1}^T \frac{\rhomax}{\rhomax}\sigma_{t-1}(x_t|\hat \Sigma_{t-1})\nonumber
   \leq \sqrt{T  \sum_{t=1}^T \frac{\rhomax^2}{k} \frac{\sigma_{t-1}^2 \big(x_t|\text{diag}(\rhomax^2/k) \big)}{ \rhomax^2/k} } \nonumber
  \\
   & \leq \sqrt{T\frac{ \rhomax^2 }{k} \frac{k/\rhomax^{2}}{\ln (1+k/\rhomax^2)} \sum_{t=1}^T \ln \bigg ( 1+  \frac{\sigma_{t-1}^2\big(x_t|\text{diag}(\rhomax^2/k) \big)}{ \rhomax^2/k} \bigg)}  \nonumber\\
   & \leq \sqrt{ \frac{2 T}{\ln (1+k/\rhomax^{2})} \underbrace{\sum_{t=1}^T \frac{1}{2}\ln \bigg ( 1+  \frac{\sigma_{t-1}^2\big(x_t|\text{diag}(\rhomax^2/k) \big)}{ \rhomax^2/k} \bigg)}_{\mathrm{mutual~ information~ \cref{eq:mutual_information_mean}}}}.
    \label{eq:bound_variances_unknown_rho}
\end{align}
Here, the first inequality follows from Cauchy-Schwarz inequality and the fact that $\sigma_t(x_t|\hat\Sigma_t) \leq \sigma_t(x_t|\text{diag}(\rhomax^2/k)).$ The latter holds by the definition of $\hat\Sigma_t$, particularly:
\begin{align}
\sigma^2_t(x_t|\hat\Sigma_t)& = \frac{1}{\lambda} (\kappa(x,x) - \kappa_t(x)^\top(K_t + \lambda \hat \Sigma_t)^{-1}\kappa_t(x)),\nonumber \\
 \sigma^2_t\big(x_t|\text{diag}(\rhomax^2/k) \big) & = \frac{1}{\lambda} (\kappa(x,x) - \kappa_t(x)^\top \big(K_t + \lambda \text{diag}(\rhomax^2/k) \big)^{-1}\kappa_t(x)),\nonumber\\
  \hat \Si_t & = \tfrac{1}{k}\text{diag}\big(\min\{\mathrm{ucb}^{\varproxy}_t(x_1), \rhomax^2\} ,\ldots, \min\{\mathrm{ucb}^{\varproxy}_t(x_t), \rhomax^2\}\big), \nonumber
\end{align}
then $\hat \Sigma_t \preceq \text{diag}(\rhomax^2/k), $ and  $-(K_t + \lambda \hat \Sigma_t)^{-1} \preceq -(K_t + \lambda \text{diag}(\rhomax^2/k))^{-1}.$ That implies 
\begin{align}
     \sigma^2_t(x_t|\hat\Sigma_t) - \sigma^2_t \big(x_t|\text{diag}(\rhomax^2/k)) &= -\kappa_t(x)^\top \big(K_t + \lambda \hat \Sigma_t)^{-1}\kappa_t(x) \big) \nonumber \\
     & + \kappa_t(x)^\top(K_t + \lambda \text{diag}(\rhomax^2/k))^{-1}\kappa_t(x) \big) \nonumber \\
    &\leq 0. \nonumber
\end{align}

The second inequality in \cref{eq:bound_variances_unknown_rho} is due to the fact that
for any $s^2\in[0,k/\rhomax^{2}(x_t)]$ we can bound $s^2 \leq \frac{k/\rhomax^{2}(x_t)}{\ln(1+k/\rhomax^{2}(x_t))} \ln(1+s^2)$, that also holds for $s^2:=\frac{\sigma_{t-1}^2\big(x_t|\text{diag}(\rhomax^2/k) \big)}{ \rhomax^2/k}$ since for any $\lambda \geq 1$ $$\frac{\sigma_{t-1}^2\big(x_t|\text{diag}(\rhomax^2/k) \big)}{ \rhomax^2/k} \leq \frac{\lambda^{-1}\kappa(x_t,x_t)}{ \rhomax^2/k} \leq k/\rhomax^{2}.$$ 

Similarly, we  bound
\begin{align}
    \sum_{t=1}^T\sigma_{t-1}^{var}(x_t) 
    & = \sum_{t=1}^T \frac{\rho_{\eta}(x_t)}{ \rho_{\eta}(x_t)}\sigma_{t-1}^{var}(x_t) \leq 
    \sqrt{T \sum_{t=1}^T  \rho^2_{\eta}(x_t) \frac{(\sigma^{var}_{t-1})^2(x_t)}{ \rho^2_{\eta}(x_t)}}\nonumber \\
   & 
     \leq 
     \sqrt{\frac{2 T}{\ln (1 + \mathcal{R}^{-2})} \underbrace{\sum_{t=1}^T \frac{1}{2}\ln \bigg( 1+ \frac{(\sigma_{t-1}^{var})^2(x_t) }{ \rho^2_{\eta}(x_t)} \bigg)}_{\mathrm{mututal~information~ \cref{eq:mutual_information_variance}}}}, \label{eq:bound_variances_unknown_rho_2}
\end{align}
in the above we define $\mathcal{R}^2 := \max_{x\in A} \rho^2_{\eta}(x), A= \{x_1, \dots, x_T\}.$

\textit{\textbf{Step 5}:} \textit{On bounding cumulative regret  $R_T=\sum_{t=1}^{T}kr(x_t)$}

Combining the above three steps together, we obtain with probability $1-2\delta$
\begin{align}
   R_T \leq \beta_T k\sqrt{\frac{2T\hat\gamma_T}{ \ln (1 + k/\rhomax^{2})}} + \alpha \beta_T^{var} k\sqrt{\frac{2T\Gamma_T }{\ln (1 + \mathcal{R}^{-2})}}. \label{eq:regret_bound_2b}
\end{align}

\subsection{Proof of \cref{col_report}}\label{corr1:proof}

\textbf{Corollary 1.1}\ \ \textit{Consider the setup of Theorem~\ref{theorem}. Let $A = \{x_t\}_{t=1}^T$ denote actions selected by \raucb over $T$ rounds. Then, with probability at least $1-\delta$, the reported point $\xreported: =  \arg\max_{x_t \in A} \lcb^{\mv}_t(x_t),$ where $ \lcb^{\mv}_t(x_t) = \mathrm{lcb}^f_{t}(x) - \alpha \;\mathrm{ucb}_{t}^{\varproxy}(x)$,  achieves $\epsilon$-accuracy, i.e., $\mv(x^*) - \mv(\xreported)\leq \epsilon$, after $T \geq\tfrac{ 4\beta_T^2\hat\gamma_T/\ln (1+   k/\rhomax^{2}) + 4\alpha (\beta_t^{var})^2\Gamma_T/\ln (1+   \mathcal{R}^{-2})}{\epsilon^2}$ rounds. }
\begin{proof}
We select the maximizer of $\lcb^{\mv}_t(x_t)$ over the past points $x_t$:
$$\hat{x}_{T}: =  x_{t^*}, \text{ where } t^* := \arg\max_t \{ \lcb^{\mv}_t(x_t)\} = \arg\min_t \{\mv(x^*) - \lcb^{\mv}_t(x_t)\},$$ 
since adding a constant does not change the solution.  We denote $ \hat r(x_t) := \mv(x^*) - \lcb^{\mv}_t(x_t). $
Then we obtain the following bound
\begin{align}\label{eq:app_mv}
    \mv(x^*) - \mv(x_{t^*}) 
    &\leq \mv(x^*) - \lcb^{\mv}_{t^*}(x_{t^*}) 
    =   \frac{1}{T}\sum_{t=1}^T \hat r(x_{t^*})\nonumber\\
    &
    \leq \frac{1}{T}\sum_{t=1}^T  \hat r(x_{t})
    = \frac{1}{T}\sum_{t=1}^T
        \left(\mv(x^*) - \lcb_t^{\mv}(x_{t})
        \right)\nonumber\\
        &
    \leq \frac{1}{T}\sum_{t=1}^T
        \left(\ucb^{\mv}_t(x^*) - \lcb_t^{\mv}(x_{t})
        \right)\nonumber
        \\
        &
        \leq \frac{1}{T}\sum_{t=1}^T
        \left(\ucb^{\mv}_t(x_t) - \lcb_t^{\mv}(x_{t})
        \right).
\end{align}
In the above,  the first inequality holds with high probability by definition  $\lcb^{\mv}_{t^*}(x_{t^*}) \leq \mv(x_{t^*}) $, the second inequality is due to $t^*:=\arg\min_t \hat r(x_t)$ and therefore $\hat r(x_{t^*}) \leq \hat r(x_t) \ \forall t=1,\ldots,T.$ 
The third inequality holds since $\ucb^{\mv}_t(x)\geq \mv(x)$ with high probability, and the fourth is due to $\ucb^{\mv}_t(x_t) \geq \ucb^{\mv}_t(x) $ for every $x$, since $x_t$ is selected via \cref{alg:risk-averse-bo}. 

Recalling \cref{eq:regret_bound_2b}, note that the following bounds hold:
\begin{align}\label{eq:app_R}
R_T 
= \sum_{t=1}^{T} k r(x_t)
\leq \sum_{t=1}^T k(\mathrm{ucb}^{\mv}_t(x_t) - \mathrm{lcb}^{\mv}_t(x_t)) 
\leq \beta_T k\sqrt{\frac{2T\hat\gamma_T}{ \ln (1 + k/\rhomax^{2})}} + \alpha \beta_T^{var} k\sqrt{\frac{2T\Gamma_T}{\ln (1 + \mathcal{R}^{-2})}. }\end{align}
 Combining the above \cref{eq:app_R} with \cref{eq:app_mv} we can get the following upper bound 
\begin{align*}
    \mv(x^*) - \mv(x_{t^*}) 
    &\leq \frac{ \beta_T k\sqrt{2T\hat\gamma_T/\ln (1+   k/\rhomax^{2})} + \alpha \beta_T^{var} k\sqrt{2T\Gamma_T/\ln(1+\mathcal{R}^{-2})}}{kT} \\
    & \leq 
     \frac{\sqrt{4 \big(k\beta_T^2\hat\gamma_T/\ln (1+   k/\rhomax^2) + \alpha k(\beta_t^{var})^2\Gamma_T/\ln (1+   \mathcal{R}^{-2}) \big)}}{\sqrt{kT}}.
\end{align*}
Therefore, for $Tk$ samples with $Tk\geq\frac{4  (k \beta_T^2\hat\gamma_T/\ln (1+   k/\rhomax^2) + \alpha k(\beta_t^{var})^2\Gamma_T/\ln (1+   \mathcal{R}^{-2}) )}{\epsilon^2}$ we finally obtain 
$$\mv(x^*) - \mv(x_{t^*}) \leq \epsilon. $$
\end{proof}


\subsection{Experimental settings and extended results}\label{app:exp_additional}

\paragraph{Implementation and resources}
We implemented all our experiments using Python and BoTorch \cite{balandat2020botorch}.\footnote{\url{https://botorch.org/}} We ran our experiments on an Intel(R) Xeon(R) CPU E5-2699 v3 @ 2.30GHz machine. 

\subsubsection{Example function }

\begin{figure}[H]
    \subfloat{
    \includegraphics[width=1.1\linewidth]{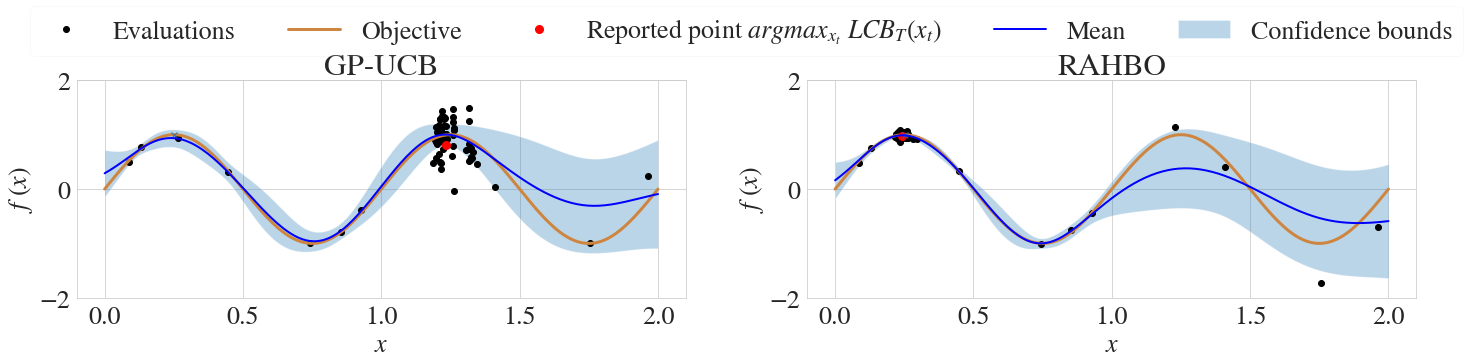}
        \label{fig:app_sine_gp2}
    }\\
    \subfloat{
    \includegraphics[width=1.1\linewidth]{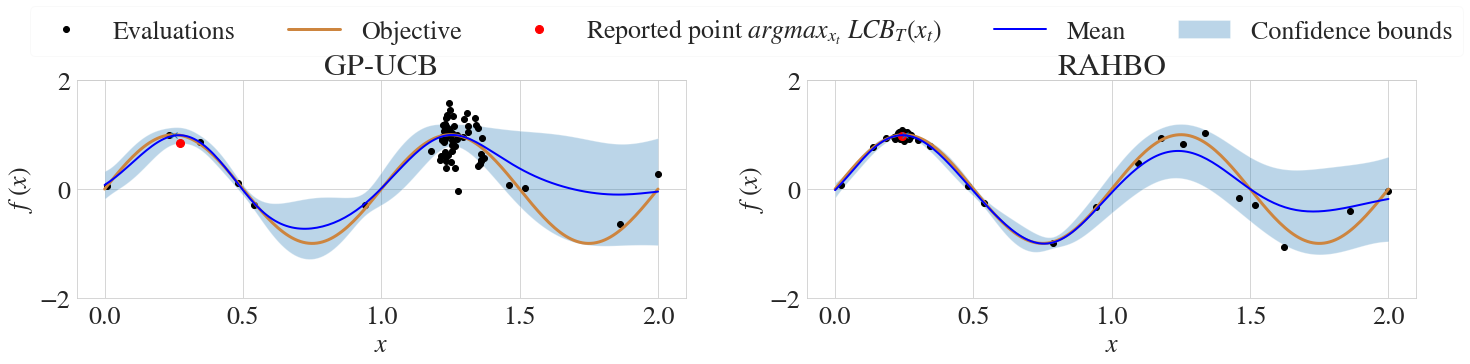}
        \label{fig:app_sine_gp3}
    }
    \caption{Additional examples for \cref{fig:sine_gp1} (each row corresponds to one initialization). GP models fitted for \gpucb  (left column) and \raucb (right column) for sine function. After  initialization with the same sampled points, \gpucb  concentrates on the high-noise region whereas \raucb prefers small variance. }
    \label{app:app_sine_gp}
\end{figure}

We provide additional visualizations for the example sine function in \cref{app:app_sine_gp}.  These examples demonstrate that exploration-exploitation trade-off (as in GP-UCB) might not be enough to prefer points with lower noise and \gpucb might tend to acquire points with higher variance. In contrast, \raucb, initialized with the same point, prefers points with lower risk inherited in noise.

\subsubsection{Branin}

\begin{figure}[H]
    \hspace*{-0.8cm}
    \subfloat[Unknown objective]{
        \includegraphics[width=0.35\linewidth]{imgs/branin/branin_abc.png}
        \label{fig:app_branin}
        }
    \hspace*{-0.5em}
    \subfloat[Unknown variance]{
        \includegraphics[width=0.35\linewidth]{imgs/branin/varince_function_abc.png}
        \label{fig:app_branin_var}
        }
    \hspace*{-0.5em}
    \subfloat[Noise-perturbed evaluations]{
        \includegraphics[width=0.35\linewidth]{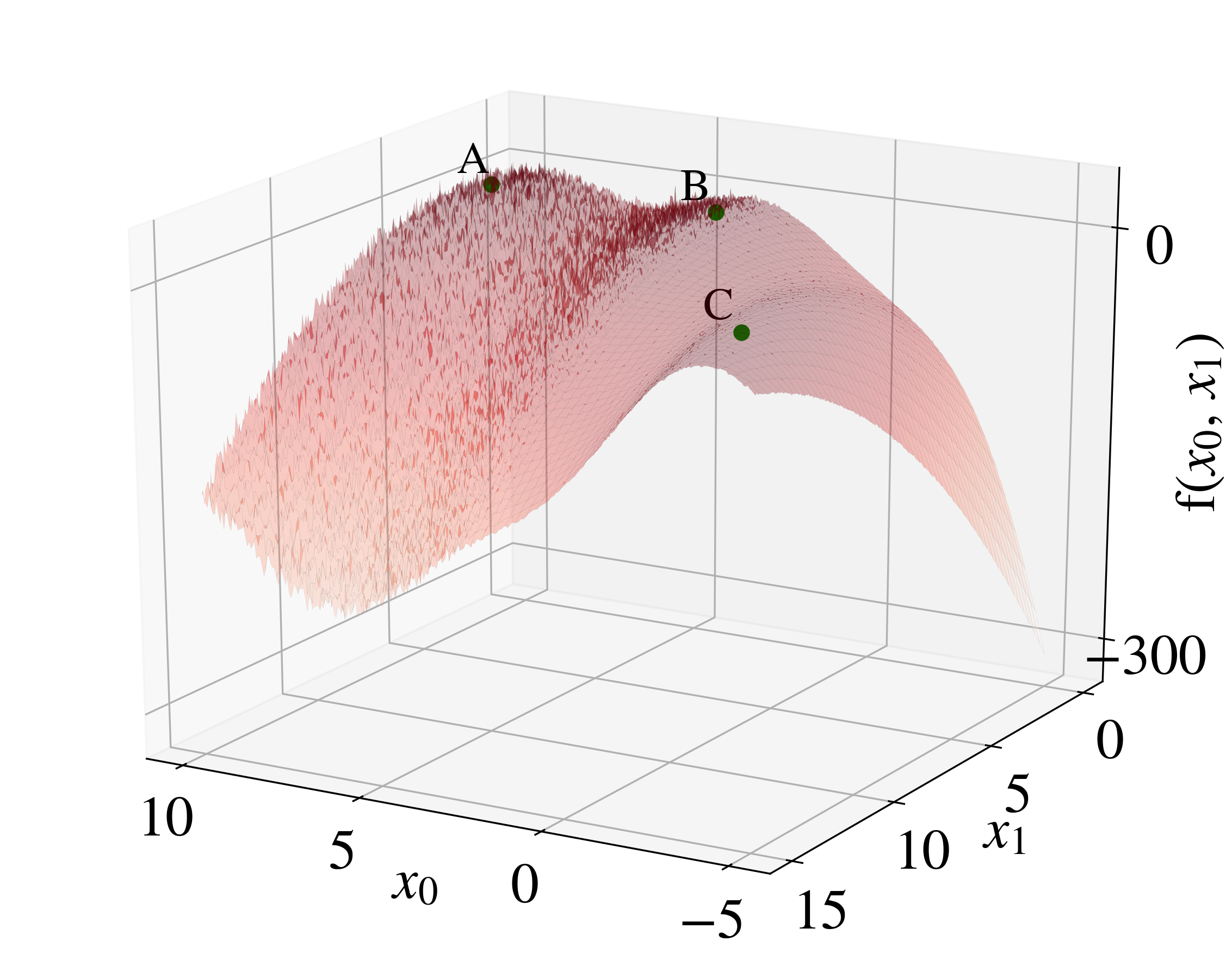}
        \label{fig:app_branin_evals}
        }
     \caption{Visualization of noise-perturbed function landscape:(a) Unknown objective with 3 global maxima marked as (A, B, C). (b) Heteroscedastic noise variance over the same domain: the noise level at (A, B, C) varies according to the sigmoid function. (c) Noise-perturbed evaluations: A is located in the noisiest region.}
     \label{fig:branin_extra}
\end{figure}

\begin{figure}[H]
    \hspace*{-0.5em}
    \subfloat[]{
        \includegraphics[width=0.35\linewidth]{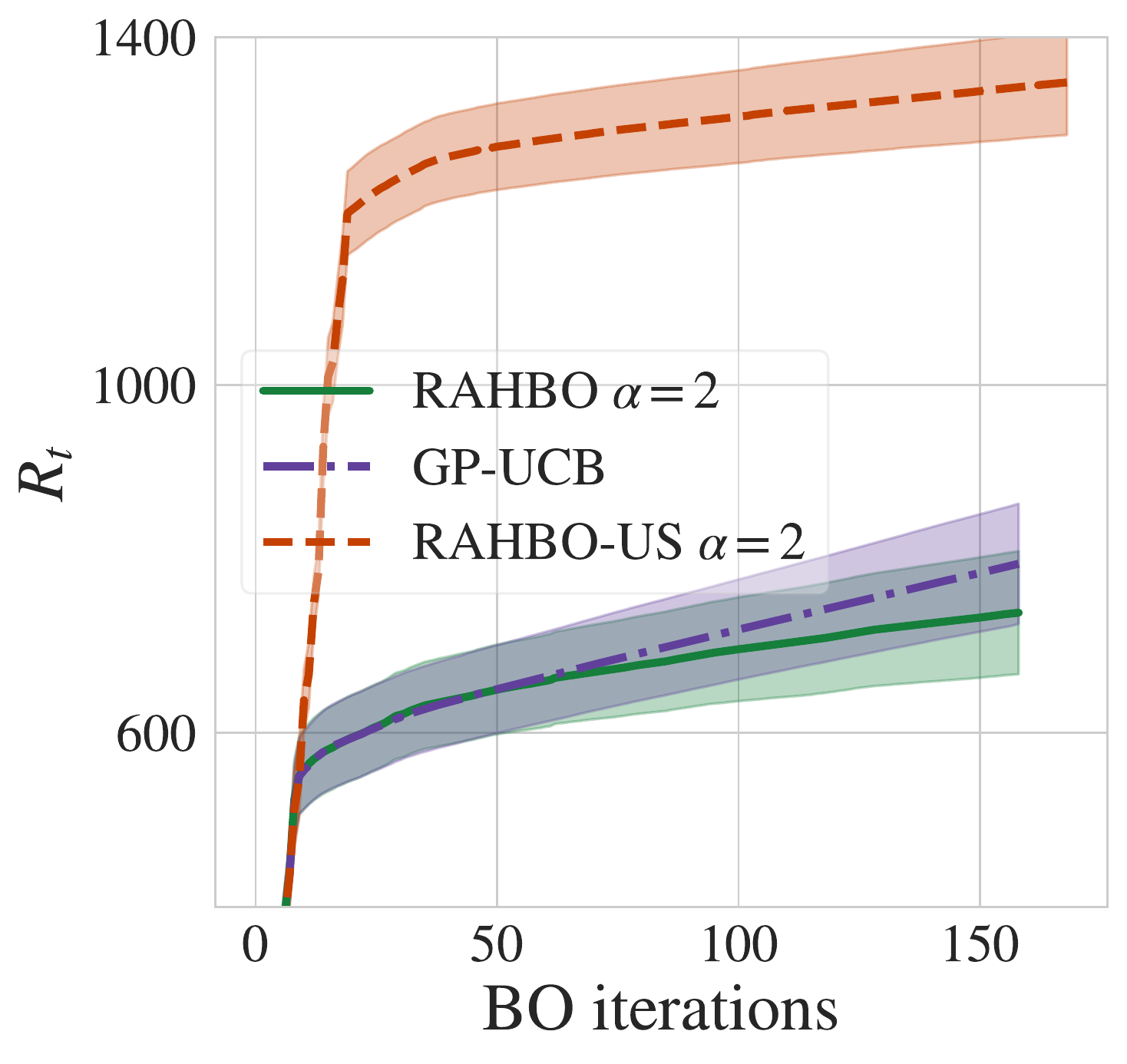}
        }
    \hspace*{-0.5em}
    \subfloat[]{
        \includegraphics[width=0.35\linewidth]{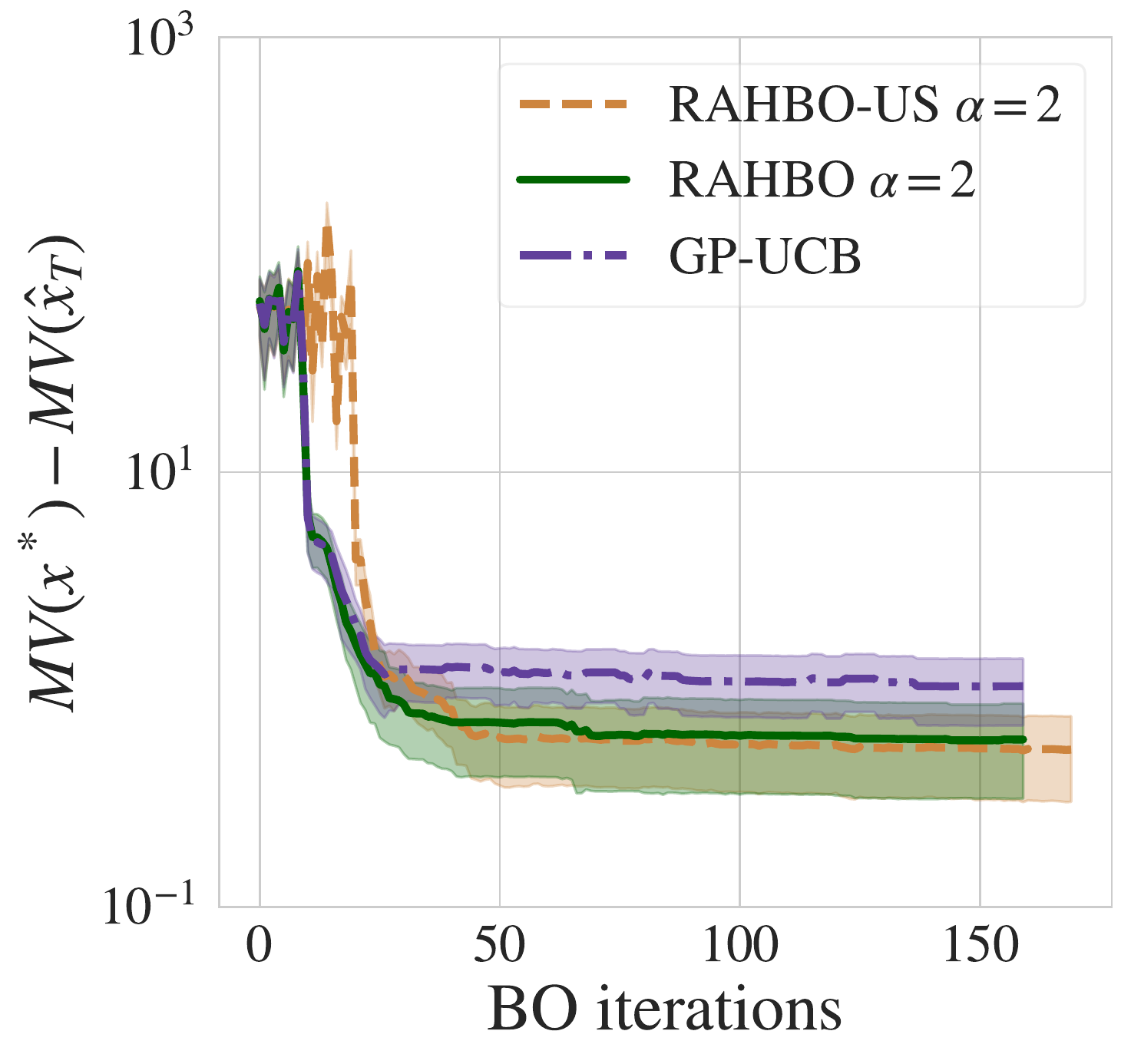}
        }
     \caption{\textbf{Branin:} (a) Cumulative regret. (b) Suboptimality w.r.t. MV}
     \label{fig:branin_regrets}
\end{figure}

We provide additional visualizations, experimental details and results. Firstly, we plot the noise-perturbed objective function in \cref{fig:branin_extra} in addition to the visualization in \cref{fig:first_figure}. In \cref{fig:branin_regrets}, we plot cumulative regret and simple mean-variance regrets that extends the results in \cref{fig:branin_cumregret} with \raucbus. The general setting is the same as described for \cref{fig:branin_cumregret}: we use $10$ initial samples, repeat each evaluation $k=10$ times, and \raucbus additionally uses $10$ samples for learning the variance function with uncertainty sampling. During the optimization, \raucbus updates the GP model for variance function after every acquired point. 
    
\subsection{Random Forest tuning}
\label{app:dataset}

\paragraph{Experiment motivation:} Consider the motivating example first: the optimized RF model will be exploited under the data drift over time, e.g., detecting fraud during a week. We are interested not only in high performance \emph{on average} but also in low variance across the results. Particularly, the first can be a realization of the decent result in the first days and unacceptable result in the last days, and the latter ensures lower dispersion over the days while keeping a reasonable mean. In this case, when training an over-parametrized model that is prone to overfitting (to the training data), e.g., Random Forest (RF) with deep trees, high variance in validation error might be observed. In contrast, a model that is less prone to overfitting can result into a similar validation error with lower variance. 

\paragraph{RF specifications:} We use scikit-learn implementation of RF. The RF search spaces for BO are listed in Table \ref{tab:parameter-range} and other parameters are the default provided by scikit-learn. \footnote{\url{https://scikit-learn.org/stable/modules/generated/sklearn.ensemble.RandomForestClassifier.html}} During BO, we transform the parameter space to the unit-cube space. 

\begin{figure}[H]
    \includegraphics[width=1\linewidth]{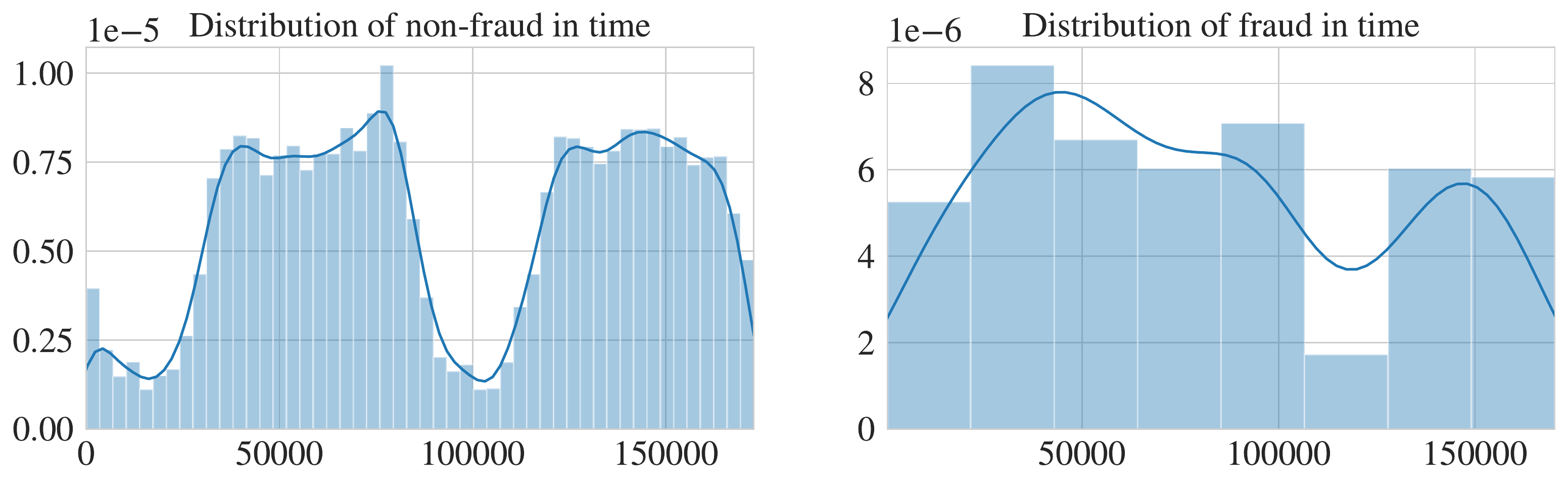}
    \caption{Distribution of non-fraud (left) and fraud (right) transactions in the dataset}
    \label{app:fraud_transactions}
\end{figure}

\paragraph{Dataset:} We tune RF on a dataset of fraudulent credit card transactions \cite{bookfrauddetection} originally announced for Kaggle competition.\footnote{\url{https://www.kaggle.com/mlg-ulb/creditcardfraud}} It is a highly imbalanced dataset that consists of ~285k transactions and only 0.2\% are fraud examples. The transactions occurred in two days and each has a time feature that contains the seconds elapsed between each transaction and the first transaction in the dataset. We use the time feature to split the data into train and validation sets such that validation transactions happen later than the training ones. The distribution of the fraud and non-fraud transactions in time is presented in \cref{app:fraud_transactions}. 

In BO, we collect evaluation in the following way: we fix the training data to be the first half of the transactions, and the rest we split into 5 validation folds that are consecutive in time. The RF model is then trained on the fixed training set, and evaluated on the validations sets. We use a balanced accuracy score that takes imbalance in the data into account. 

\begin{table}[h]
    \caption{Search space description for RF.}
     \label{tab:parameter-range}
    \begin{adjustbox}{width=0.5\columnwidth,center}
        \begin{tabular}{lrr}
            \toprule
            task &  hyperparameter & search space \\
            \midrule
             \multirow{3}{*}{RandomForest}&{n\_estimators}&{[$1$, $100$]}\\
             &{max\_features}&[{$5$, $28$]}\\
             &{max\_depth}&{[$1$, $15$]}\\
        \bottomrule
        \end{tabular}
    \end{adjustbox}
\end{table}

\subsubsection{Tuning Swiss free-electron laser (\swissfel)} \label{app:swissfel}


\begin{figure}[h]
 \hspace{-1cm}
    \subfloat[Mean-variance tradeoff (\swissfel)]{
    \includegraphics[width=0.48\linewidth]{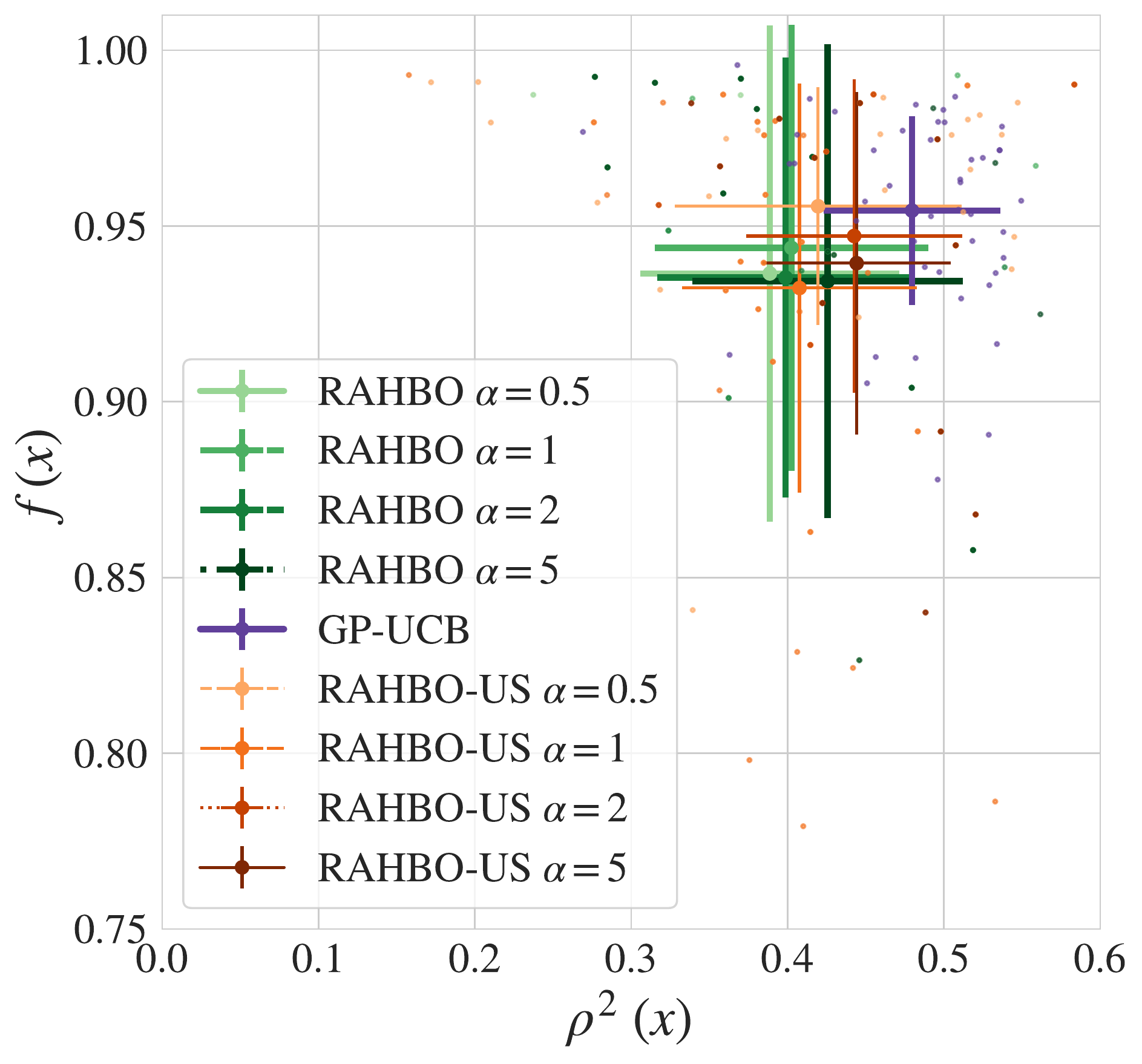}
    \label{fig:swissdel_reported_app}
    }
    \subfloat[Cum. regret ($\alpha=0.5$)]{
    \includegraphics[width=0.26\linewidth]{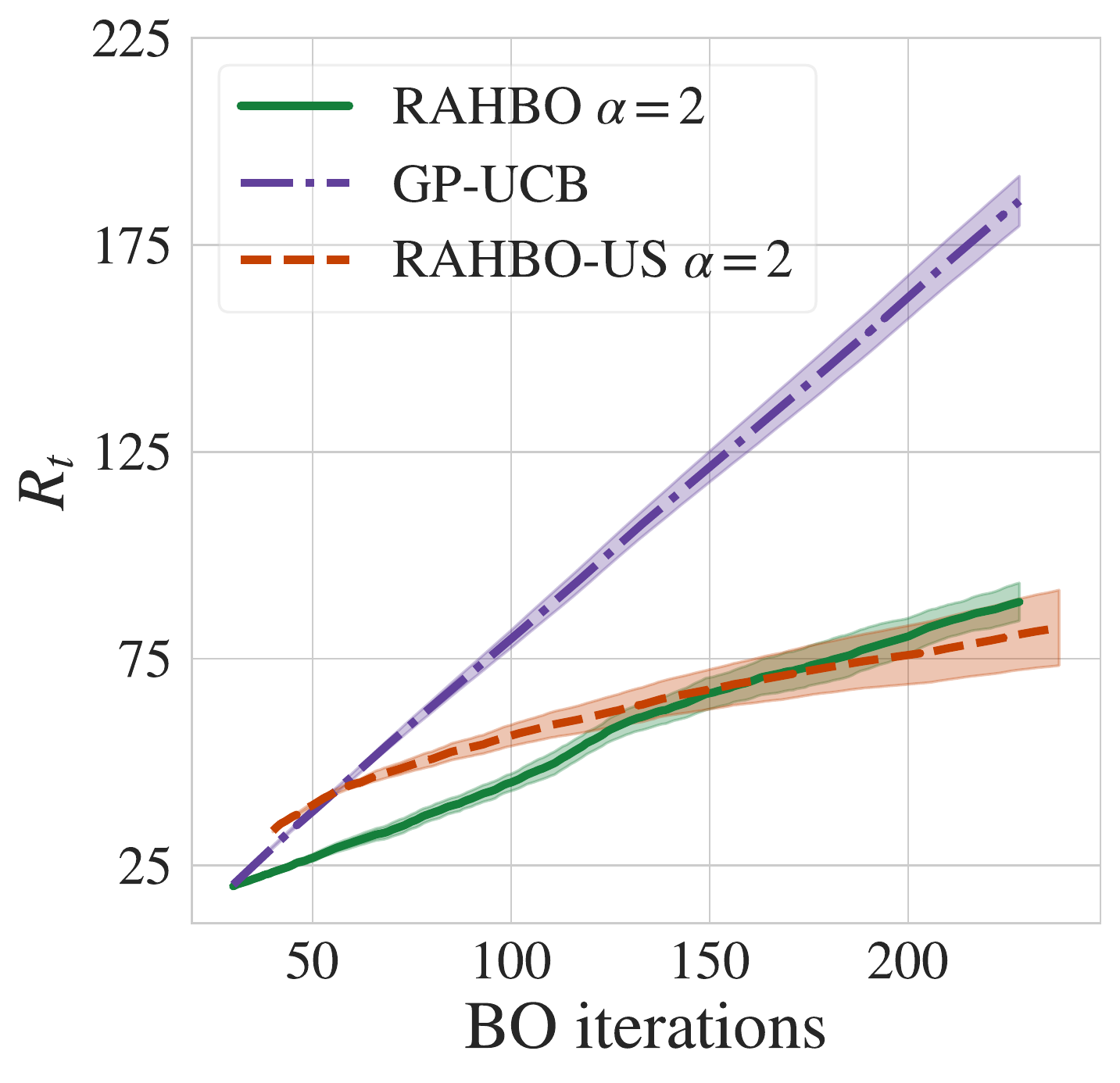}
    \label{fig:swissfel_cumregret_a_app}
    }
  \subfloat[Cum. regret ($\alpha=1$)]{
    \includegraphics[width=0.26\linewidth]{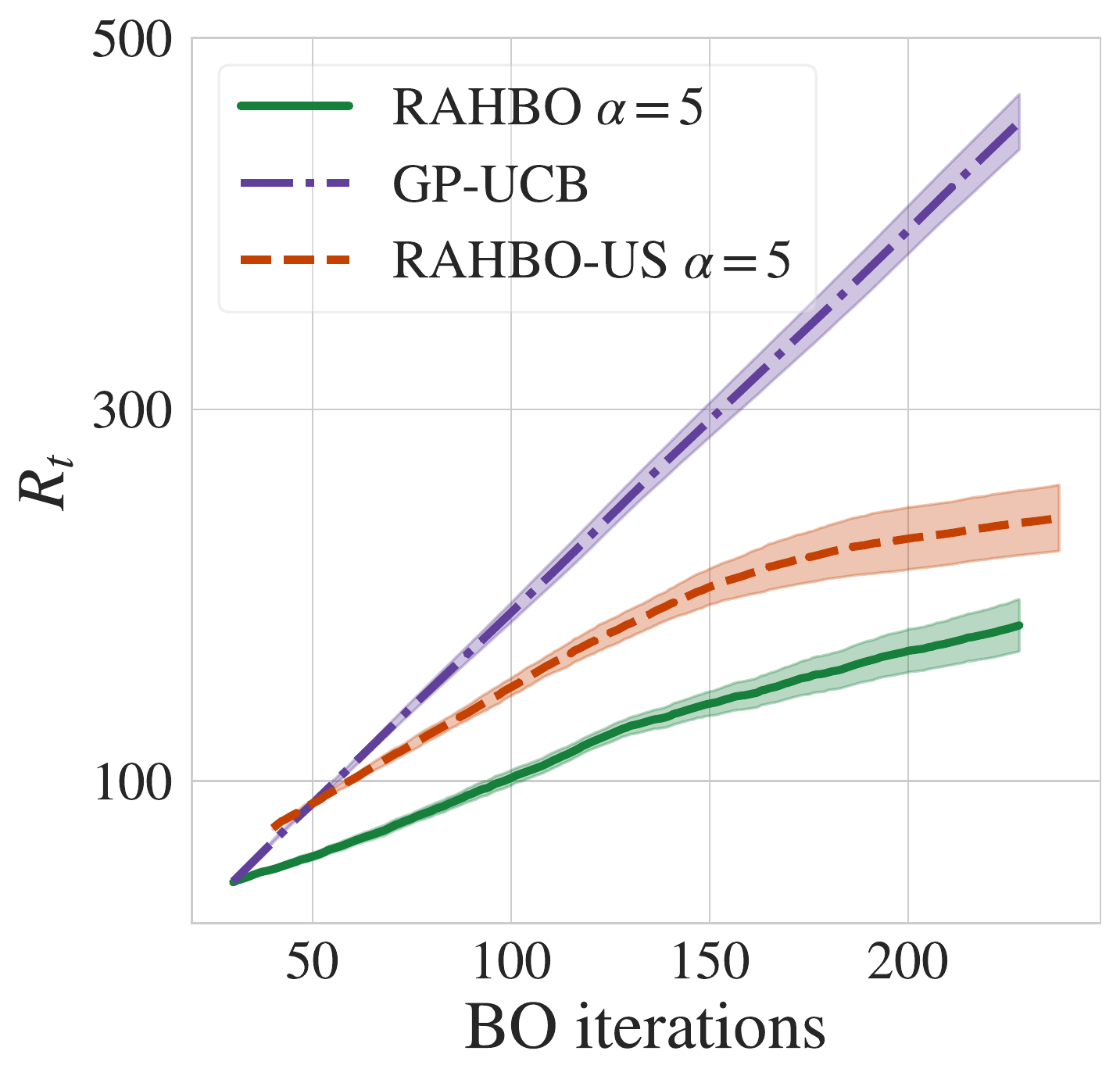}
    \label{fig:swissfel_cumregret_b_app}
    }
\caption{(a) We plot standard deviation error bars for $f(x)$ and $\rho^2(x)$ at the \textit{reported point} by the best observed value $x^{(T)} = \argmax_{x_t} y_t(x_t)$ after BO completion for \swissfel. The mean and std of the error bars are taken over the repeated BO experiments. The results demonstrate, that reporting based on the best observed value inherits high noise and as the result all methods perform similarly. Intuitively, when noise variance is high, it is possible to observe higher values. That however also inherits observing much lower value at this point, this leading to very non-robust solutions. (b-c) Cumulative regret for $\alpha  = 2$ and $\alpha = 5$.}
\label{fig:app_swissfel_res}
\end{figure}
\end{document}